\documentclass[acmsmall]{acmart}

\AtBeginDocument{%
  }

\acmDOI{10.1145/3617827}

\acmJournal{JACM}
\acmVolume{42}
\acmNumber{2}
\acmArticle{47}
\acmMonth{11}
\acmYear{2023}

\usepackage{amsmath}

\usepackage{multirow}
\usepackage{enumitem}
\usepackage{amssymb}
\usepackage{graphicx}
\usepackage{stmaryrd}
\usepackage{color}
\usepackage{xcolor}
\begin{document}

\title{Dynamic Multimodal Fusion via Meta-Learning Towards Micro-Video Recommendation}




\author{Han Liu}
\affiliation{%
  \institution{Shandong University}
  \streetaddress{School of Computer Science and Technology}
  \city{Qingdao}
  \country{China}
}
\email{hanliu.sdu@gmail.com}

\author{Yinwei Wei}
\affiliation{%
 \institution{National University of Singapore}
 \streetaddress{School of Computing}
 \country{Singapore}}
\email{weiyinwei@hotmail.com}

\author{Fan Liu}
\affiliation{%
 \institution{National University of Singapore}
 \streetaddress{School of Computing}
 \country{Singapore}}
\email{liufancs@gmail.com}

\author{Wenjie Wang}
\affiliation{%
 \institution{National University of Singapore}
 \streetaddress{School of Computing}
 \country{Singapore}}
\email{wenjiewang96@gmail.com}

\author{Liqiang Nie}
\authornote{Corresponding author}
\affiliation{%
  \institution{Harbin Institute of Technology (Shenzhen)}
  \streetaddress{School of Computer Science and Technology}
  \city{Shenzhen}
  \country{China}}
\email{nieliqiang@gmail.com}

\author{Tat-Seng Chua}
\affiliation{%
  \institution{National University of Singapore}
  \streetaddress{School of Computing}
  \country{Singapore}
  }
\email{dcscts@nus.edu.sg}



\renewcommand{\shortauthors}{Liu et al.}

\begin{abstract}
Multimodal information (\textit{e.g.}, visual, acoustic, and textual) has been widely used to enhance representation learning for micro-video recommendation. For integrating multimodal information into a joint representation of micro-video, multimodal fusion plays a vital role in the existing micro-video recommendation approaches. However, the static multimodal fusion used in previous studies is insufficient to model the various relationships among multimodal information of different micro-videos. In this paper, we develop a novel meta-learning-based multimodal fusion framework called \textit{Meta Multimodal Fusion} (MetaMMF), which dynamically assigns parameters to the multimodal fusion function for each micro-video during its representation learning. Specifically, MetaMMF regards the multimodal fusion of each micro-video as an independent task. Based on the meta information extracted from the multimodal features of the input task, MetaMMF parameterizes a neural network as the item-specific fusion function via a meta learner. We perform extensive experiments on three benchmark datasets, demonstrating the significant improvements over several state-of-the-art multimodal recommendation models, like MMGCN, LATTICE, and InvRL. Furthermore, we lighten our model by adopting canonical polyadic decomposition to improve the training efficiency, and validate its effectiveness through experimental results. Codes are available at https://github.com/hanliu95/MetaMMF.
\end{abstract}



\maketitle
\section{Introduction}
Micro-videos have become the emerging mediums of recording and sharing people's lives, due to their rich expression capacity via multimodal information, involving visual, acoustic, and textual modalities. Nowadays, popular micro-video platforms, such as Tiktok and Kwai, are exposed to an influx of users and micro-videos, attracting increasing research attention on micro-video recommendation~\cite{jiang2020aspect,cai2021heterogeneous,ma2018lga}. 

Multimodal information (\textit{e.g.}, frames, audios, and captions) associated with items\footnote{Without special instructions, items refer to micro-videos indiscriminately in this paper.} has been widely used for representation learning in micro-video recommendation because of its capability to alleviate the user-item interaction sparsity problem~\cite{chen2017attentive,du2020learn}. In other words, recommender systems can more comprehensively understand user preferences and item characters by incorporating multimodal information for representation learning.
Existing multimodal recommendation models integrate multimodal information into a joint multimodal representation by employing various multimodal fusion methods. We roughly classify existing multimodal fusion into two categories: linear~\cite{wei2020graph} and nonlinear~\cite{du2020learn} methods. For example, MMGCN~\cite{wei2019mmgcn} performs the graph convolutional operations on user-item graphs in different modalities, and implements multimodal fusion via the linear combination of multimodal representations. Due to the limitations of linear methods, deep neural networks are employed to fuse multimodal information nonlinearly. For instance, VBPR~\cite{he2016vbpr} adopts a neural layer to project the multi-modality features into the latent space. ACF~\cite{chen2017attentive} introduces a hierarchical attention mechanism to select informative content by assigning weights to different modalities.  

\begin{figure}[tbp]
\centering
\includegraphics[width=\linewidth]{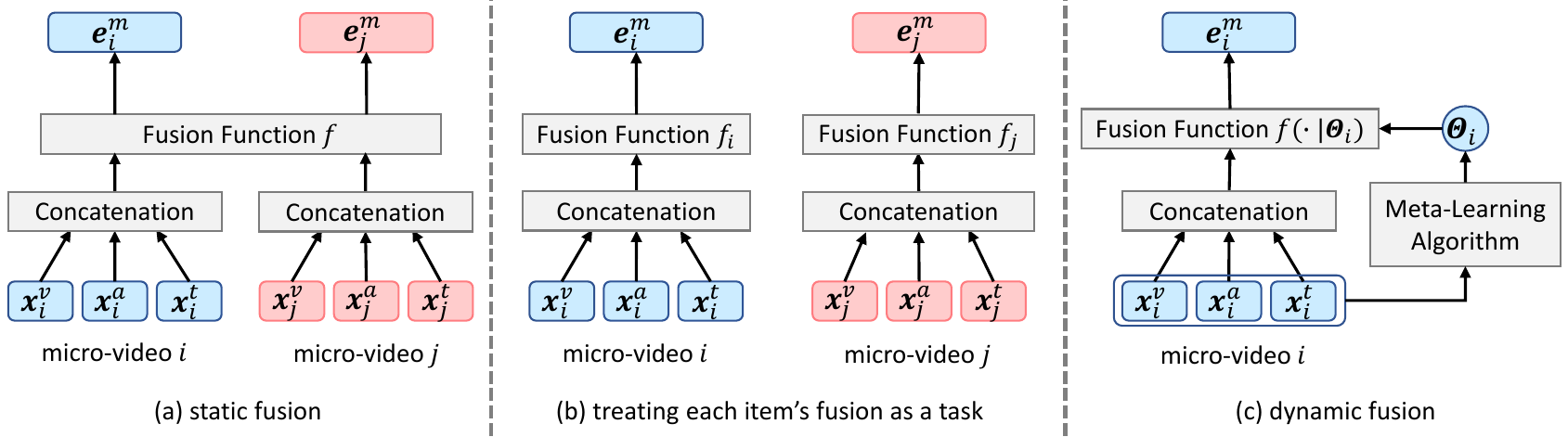}
\caption{Illustration of the static and the dynamic multimodal fusion. $\mathbf{x}_i^v$, $\mathbf{x}_i^a$, and $\mathbf{x}_i^t$ denote the features derived from visual, acoustic, and textual modalities of micro-video $i$, respectively. $\mathbf{e}_i^m$ denotes the multimodal representation of micro-video $i$ through the fusion function. $\Theta_i$ denotes the item-specific parameters dynamically generated for the fusion function by a meta-learning algorithm. Similar notations are used for denoting the attributes of micro-video $j$.}
\label{fig:introduction}
\vspace{-0.5cm}
\end{figure}

Despite the remarkable performance, the previous methods are only able to perform static multimodal fusion as shown in Figure~\ref{fig:introduction}(a), following the assumptions that: 1) for different items, the information from the same modality plays a similar role and the multimodal information obeys the same relationship; and 2) a fusion function can be learned from the vast amount of items, and that the function learned is capable of modeling the relationships among the multimodal information. However, we argue that the above-mentioned assumptions are fragile in micro-video recommendation. In particular, unlike the well-designed video (\textit{e.g.}, movies or documentaries), the modality-specific information of different micro-videos may have very different effects on the expression of the theme, and the relationships among modalities might also be distinct~\cite{baltruvsaitis2018multimodal}. For example, the information in visual and acoustic modalities is complementary in music micro-videos; while in education micro-videos, acoustic and textual modalities convey the themes in a consistent relationship. With this consideration, we argue that static fusion is under-fitting the varying multimodal fusion for learning representations of micro-videos, leading to suboptimal recommendation performance. Furthermore, we consider that each micro-video should combine its visual, acoustic, and textual modality information in a unique way. 

In this paper, we propose a dynamic multimodal fusion method that treats the multimodal fusion of each item as an independent task. As shown in Figure~\ref{fig:introduction}(b), the item-specific fusion functions are learned for different items independently. For implementation, we resort to the meta-learning techniques to establish the dynamic multimodal fusion as shown in Figure~\ref{fig:introduction}(c), due to their following advantages:
\begin{itemize}
    \item Meta-learning algorithm is capable of generating adaptive parameters according to the context of a single task~\cite{zhu2021learning, fu2020depth}, which is consistent with the target of dynamic multimodal fusion strategy.
    \item Meta-learning algorithm offers an effective way to perform transfer learning across tasks by means of shared parameters~\cite{vartak2017meta}, thus enabling us to effectively learn the item-specific fusion functions with the limited training samples.   
\end{itemize}
 
More specifically, we develop a novel meta-learning-based multimodal fusion model, named Meta Multimodal Fusion (MetaMMF), to dynamically integrate multimodal information for micro-video recommendation. In particular, MetaMMF consists of two main components: a meta information extractor and a meta fusion learner. The former is implemented by a multi-layer perceptron (MLP). It projects the multimodal features of the input task (for an item) into the meta information that represents the task features in the form of higher-order abstraction~\cite{munkhdalai2017meta}. Based on the meta information, the meta fusion learner parameterizes a neural network as the fusion function for the input task. Namely, the meta fusion learner establishes a mapping from meta information to fusion parameters, while generalizing the learning-to-learn (meta-learning) mechanism to tasks. To parameterize a layer in the neural network, the meta fusion learner exploits a task-shared 3-D tensor to transform the meta information vector into the layer's weight matrix by multiplication. Mathematically, by regarding the 3-D tensor as a set of 2-D matrices, the transformation is actually a linear combination of these 2-D matrices weighted by the elements in meta information, thus satisfying the dynamic nature of the results. Moreover, considering the model complexity problem caused by excessive parameters, we adopt \textit{Canonical Polyadic} (CP) Decomposition to lighten our model by decomposing the 3-D tensors into fewer parameters. In this way, MetaMMF can dynamically learn an item-specific fusion function for each item to integrate its multimodal features into the multimodal representation. After combining with the collaborative representation, the joint representation can then be directly used to predict the interactions between the users and micro-videos like MF or passed through an advanced model like GCN. We conduct extensive experiments on three publicly accessible datasets to verify the rationality and effectiveness of our MetaMMF method.

Overall, the three main contributions of our work are summarized as follows.
\begin{itemize}[leftmargin=*]
\item We highlight the importance of dynamic multimodal fusion for effective representation learning in micro-video recommendation. Towards this end, we present a meta-learning-based method to fuse multimodal information of each item dynamically. This work represents one of the pioneering efforts in utilizing meta-learning for dynamic multimodal fusion.
\item We devise a model MetaMMF for dynamic multimodal fusion. By analyzing the to-be-fused multimodal features of an item, MetaMMF dynamically parameterizes a neural network utilized as the item-specific fusion function. As a technical contribution, we innovatively employ tensor decomposition to downsize the model parameters, resulting in a lighter model with improved training efficiency.  
\item To validate the effectiveness of our proposed method, we conducted extensive experiments on three real-world datasets. The experimental results demonstrate that MetaMMF achieves state-of-the-art performance in enhancing multimodal representation learning.
\end{itemize}

\section{Related Work}
In this section, we review existing work on micro-video recommendation, multimodal fusion, and meta-learning, which are the most relevant to this work.

\subsection{Micro-Video Recommendation}
Consistent with the general personalized recommendation,  the studies in this field parameterize users and micro-videos with embeddings, and reconstruct their interactions to learn parameters. Recently, multiple modalities of micro-videos are widely exploited to enhance their representations in related literature because they can provide rich auxiliary information~\cite{liu2018MAML}. 

One of the early studies, Visual Bayesian Personalized Ranking (VBPR)~\cite{he2016vbpr} model, projects the modal features of each micro-video into a modal representation, then concatenated with the collaborative representation to form the final item representation. Such a fusion can only capture a linear relationship. With the rise of deep learning, many studies have successfully employed the non-linear fusion of multimodal features. For example, User-Video Co-Attention Network (UVCAN)~\cite{liu2019user} utilizes attention mechanisms at the user and micro-video levels to selectively incorporate modality information for representation learning. Nevertheless, its fusion method, a weighted sum of multimodal features, results in coarse-grained multimodal relationships. More recently, Graph Convolution Network (GCN)~\cite{hamilton2017inductive} has gained popularity for recommendation representation learning due to its outstanding performance. MMGCN~\cite{wei2019mmgcn} builds modality-aware GCNs based on a user-item bipartite graph to learn unimodal representations for users and micro-videos. However, its final multimodal fusion relies on a linear combination, neglecting the varying impact of different modalities. Heterogeneous Hierarchical Feature Aggregation Network (HHFAN)~\cite{cai2021heterogeneous} leverages a modality-aware heterogeneous information graph to explore complex relationships among users, micro-videos, and multimodal information, generating high-quality representations. Nevertheless, the fusion of the neighbor's multimodal features in HHFAN is performed using a static aggregation network. Invariant Representation Learning (InvRL)~\cite{du2022invariant} learns invariant representations that account for user attention and mitigate the influence of spurious correlations in micro-video recommendations. The learned invariant representations can consistently predict user-item interactions across different environments. Despite this innovation, its fusion module simply concatenates multimodal features.

Although these previous studies have shown performance improvements, their static fusion approach limits the capture of diverse multimodal relationships in micro-videos, resulting in suboptimal representations and negatively affecting recommendation performance. To address this limitation, we propose a novel dynamic multimodal fusion-based recommendation model, enhancing the representation learning of micro-video by dynamically fusing multimodal information. Our model utilizes the multimodal features of micro-videos and leverages meta-learning to dynamically parameterize a fusion neural network for each micro-video. This adaptive approach enables the capture of specific multimodal relationships pertaining to each micro-video. Furthermore, our model is designed to be model-agnostic and lightweight, providing distinct advantages in flexibility and efficiency compared to previous methods.

\subsection{Multimodal Fusion}
Multimodal fusion is one of the traditional topics in multimodal machine learning. In technical terms, multimodal fusion refers to integrating information from multiple modalities to predict a class through classification or a continuous value through regression~\cite{baltruvsaitis2018multimodal}. Most studies on multimodal fusion are model-agnostic, which can be divided into early (\textit{i.e.}, feature-based), late (\textit{i.e.}, decision-based), and hybrid fusion.

Early fusion integrates features extracted from different modalities. For example, ACNet~\cite{hu2019acnet} utilizes the attention mechanism to integrate valuable features from RGB and depth branches adaptively. However, the fusion process only considers a complementary relationship and utilizes a static layer with element-wise addition. In contrast, late fusion combines the output results of multiple models for multiple modalities. The related studies focus on fusing unimodal decision values, such as the neural multimodal cooperative learning (NMCL)~\cite{wei2019neural} model, which enhances individual features in each modality using cooperative nets and performs late fusion on the prediction results from different modalities. However, these approaches overlook the feature-level interactions between modalities. Hybrid fusion, on the other hand, combines outputs from early fusion and individual unimodal predictors, aiming to leverage the advantages of both above-described methods in a common framework, such as Combine Early and Late Fusion Together (CELFT)~\cite{wang2021combine}. This universal hybrid framework considers the fusion of different levels regardless of the fusion dynamics.

These fusion approaches adhere to an immutable formulation, \textit{i.e.}, the fusion process remains constant regardless of the input. Such fusion strategies are unaware of the multimodal content of items. Considering that the relationships among modalities vary with the multimodal features of each item, we introduce a meta-learning-based multimodal fusion method. Our method dynamically generates item-specific fusion parameters by analyzing the input multimodal features, enabling us to achieve a superior fusion capability.

\subsection{Meta-Learning in Recommendation}
Meta-learning, also known as learning-to-learn, is inspired by the human learning process, which can quickly learn new tasks based on a small number of examples. Meta-learning tends to train a model that can adapt to a new task that is not used during the training with a few examples~\cite{lee2019melu,cui2021sequential}. The related studies can be classified into three types: metric-based~\cite{snell2017prototypical}, memory-based~\cite{santoro2016meta}, and optimization-based meta-learning~\cite{finn2017model}. 

Metric-based methods learn a metric or distance function over tasks, while model-based methods aim to design an architecture or training process for rapid generalization across tasks. Lastly, optimization-based methods directly adjust the optimization algorithm to enable quick adaptation with just a few examples. Previous studies~\cite{snell2017prototypical,santoro2016meta} on metric-based and memory-based meta-learning convert the recommendation task to the classification problem. The work in~\cite{vartak2017meta} implements a meta-learning strategy by learning a neural network classifier whose biases are determined by the item history. The classifier performs recommendations by predicting whether a user consumes an item. A series of recent studies~\cite{zhu2021learning} have adopted the optimization-based meta-learning approach and chose model-agnostic meta-learning (MAML)~\cite{finn2017model} for model training. For example, MeLU~\cite{lee2019melu} applies the MAML framework to score the affinities, which rapidly adapts to new users or items based on sparse interaction history.

From the perspective of meta-learning theory, learning the multimodal fusion process of each micro-video can be regarded as an individual task with limited training samples. Inspired by this, we propose a meta-learning strategy to implement dynamic multimodal fusion for learning robust representation of each micro-video in the recommendation scenario.
\section{Preliminaries}
In this section, we first introduce the task of micro-video recommendation. We then shortly recapitulate the widely used multimodal representation learning based on neural networks, highlighting the limitation caused by using static multimodal fusion. Finally, we formalize the problem of dynamic multimodal fusion.  For ease of reading, we use a bold uppercase letter to denote a matrix, a bold lowercase letter to denote a vector, an italic letter to denote a scalar, and a calligraphic uppercase letter to denote a set.

\subsection{Micro-Video Recommendation}
Micro-video recommendation focuses on inferring the preference degree of user $u$ on micro-video item $i$. Basically, all the models can be composed of two parts: representation learning and interaction prediction~\cite{liu2018discrete}. The former is responsible for yielding vector representations $\mathbf{e}_u$ and $\mathbf{e}_i$ for user $u$ and item $i$, respectively. Then the preference score of user $u$ on item $i$ can be estimated with the function,
\begin{equation}
    \hat{y}_{ui} = \rho(\mathbf{e}_u, \mathbf{e}_i),
\end{equation}
where $\rho(\cdot)$ denotes the prediction function (\textit{e.g.}, inner product). Since each item is associated with multimodal (\textit{e.g.}, visual, acoustic, and textual) features, many efforts have been made to exploit them to enhance the item representation. In recent studies, a typical operation is to represent item $i$ by concatenating its collaborative representation and multimodal representation,
\begin{equation}\label{eq:cat}
    \mathbf{e}_i = \begin{bmatrix}\mathbf{e}_i^m\\\mathbf{e}_i^c\end{bmatrix},
\end{equation}
where $\mathbf{e}_i^c\in\mathbb{R}^{d_c}$ indicates the collaborative representation bound up with ID, and $\mathbf{e}_i^m\in\mathbb{R}^{d_m}$ indicates the multimodal representation learned by integrating features from multiple modalities. $d_c$ and $d_m$ denote the dimensions of the two kinds of representation, respectively.

\subsection{Multimodal Fusion for Multimodal Representation Learning}
In order to obtain multimodal representation, a procedure is required to integrate multimodal features into the same embedding space~\cite{liu2022disentangled}. Multimodal fusion emerges to resolve how to combine the information from heterogeneous sources and represent information in a format that a computation model can work with. Therefore, it is widely applied to learning joint multimodal representations in micro-video recommendation. For item $i$, its multimodal representation $\mathbf{e}_i^m$ can be mathematically expressed as,
\begin{equation}
    \mathbf{e}_i^m = f(\mathbf{x}_i^v, \mathbf{x}_i^a, \mathbf{x}_i^t),
\end{equation}
where $\mathbf{x}_i^v\in\mathbb{R}^{d_v}$, $\mathbf{x}_i^a\in\mathbb{R}^{d_a}$, and $\mathbf{x}_i^t\in\mathbb{R}^{d_t}$ denote the features deriving from visual, acoustic, and textual modalities of item $i$, respectively. $d_v$, $d_a$, and $d_t$ denote the dimensions of these modality features, respectively. $f(\cdot)$ denotes the multimodal fusion function. Considering the ability to learn complex decision boundaries, recent studies have widely adopted neural networks as the multimodal fusion function~\cite{mroueh2015deep,ngiam2011multimodal}. For example, VBPR uses one neural layer to implement multimodal fusion by projecting the concatenation of the visual, acoustic, and textual modality features,
\begin{equation}
\mathbf{e}_i^m=\mathbf{W}\begin{bmatrix}\mathbf{x}_i^v\\\mathbf{x}_i^a\\\mathbf{x}_i^t\end{bmatrix},
\end{equation}
where $\mathbf{W}\in\mathbb{R}^{d_m\times(d_v+d_a+d_t)}$ denotes the weight matrix parameter. However, such multimodal fusion functions with static parameters cannot model the various relationships among multiple modalities in different micro-videos.

\subsection{Problem Set-Up of Dynamic Multimodal Fusion}
We focus on the dynamic multimodal fusion problem in micro-video recommendation. In this paper, static fusion regards the multimodal fusion of all items as the same task. In contrast,  we treat multimodal fusion for each item as an individual task in dynamic fusion. We aim to generate the item-specific parameters in a neural network format for each fusion task to achieve dynamic multimodal fusion. As discussed in Section~1, meta-learning is a potential solution due to its ability to learn parameters for multiple tasks and address the few-sample training problem for each task. In particular, we aim to develop a meta-learning model capable of processing an item's multimodal features as input, and generating neural network parameters that can be utilized to fuse the different modalities of the item. Let $\mathcal{X}_i=\{\mathbf{x}_i^v, \mathbf{x}_i^a, \mathbf{x}_i^t\}$ be the set of multimodal features belonging to item $i$. The learning task of item $i$ is to obtain its multimodal representation as the output of a neural network fusion function $f(\mathcal{X}_i|\Theta_i)$ where the parameters $\Theta_i$ are produced from the multimodal features to be fused:
\begin{equation}\label{eq:problem}
    \mathbf{e}_i^m=f(\mathcal{X}_i|\mathcal{G}(\mathcal{X}_i)).
\end{equation}
The meta-learning refers to learning the function $\mathcal{G}(\mathcal{X}_i)$ that takes multimodal features $\mathcal{X}_i$ as input and produces the parameters of the neural network $f(\mathcal{X}_i|\Theta_i)$. In the following section, we describe the framework for learning $\mathcal{G}(\mathcal{X}_i)$ in detail.
\section{Method}
\begin{figure*}[tbp]
\centering
\includegraphics[width=0.53\linewidth]{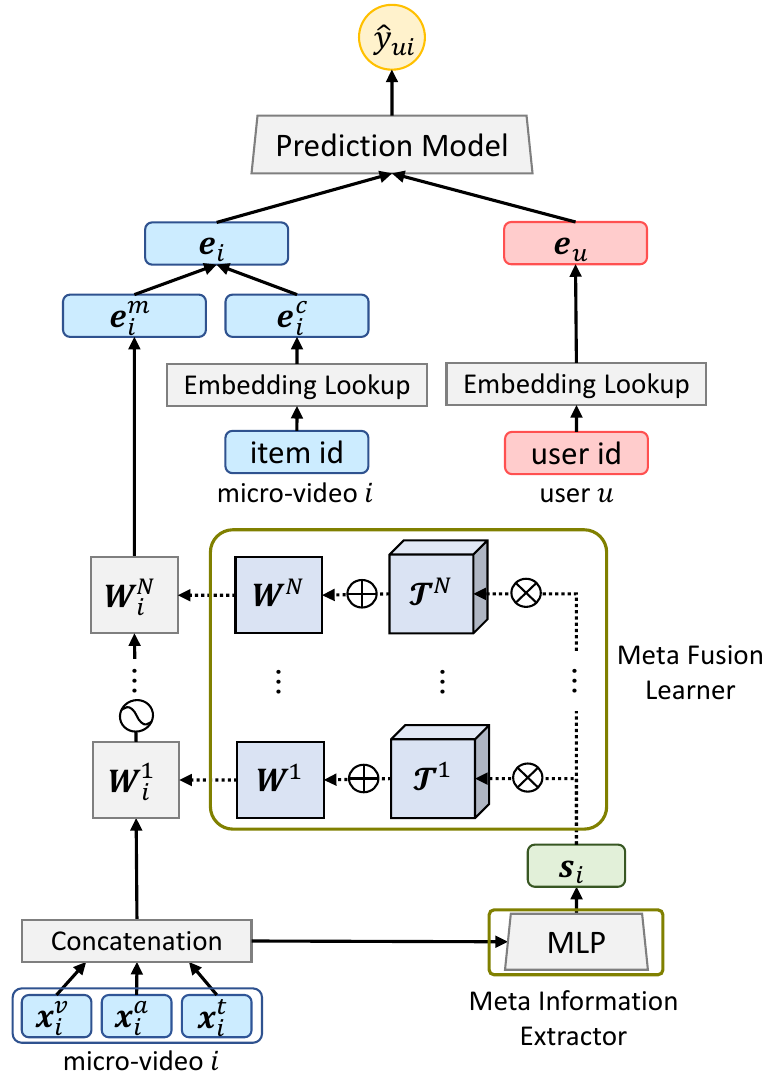}
\caption{Schematic illustration of our proposed model.}
\label{fig:framework}
\end{figure*}

We first present the general MetaMMF framework as illustrated in Figure~\ref{fig:framework}, elaborating on how to learn to parameterize a neural layer for fusing the multimodal features of a micro-video. Based on this, we then extend the single-layer structure into a multi-layer one. To lighten the space complexity of MetaMMF, we present a parameter simplification method based on tensor decomposition. Furthermore, we show that MetaMMF is model-agnostic and can be used as the early step prior to other representation learning functions, such as MF and GCN. Lastly, we provide the optimization of MetaMMF.

\subsection{General One-Layer Fusion Framework}
As discussed in Section 1, we view dynamic multimodal fusion from a meta-learning perspective, and regard the multimodal fusion of each micro-video as an individual learning task. Formulated as Eqn.(\ref{eq:problem}), we propose a meta-learning model MetaMMF to learn a neural network for each task in terms of the micro-video's multimodal information, and thus yield an item-specific multimodal fusion function for representation learning. We first illustrate how MetaMMF dynamically learns the parameter of a neural layer, and then generalizes it to a multi-layer network.

Specifically, a meta-learning function $\mathcal{G}$ is responsible for generating the adaptive parameters of each fusion task in a neural network format by processing the meta information. Considering that multimodal features can reflect the relationships among modalities in micro-videos, we employ them as the source of meta information to decide their fusion parameters. In this way, the dynamic parameter generation of a one-layer fusion framework is defined as: 
\begin{equation}
    \mathbf{W}_i = \mathcal{G}(\mathcal{X}_i),
\end{equation}
where $\mathbf{W}_i\in\mathbb{R}^{d_m\times(d_v+d_a+d_t)}$ is the weight matrix of a neural layer specific to item $i$, produced by the meta-learning function $\mathcal{G}$ from the multimodal features of item $i$. In particular, the design of function $\mathcal{G}$ is twofold, including meta information extractor and meta fusion learner.

\subsubsection{Meta Information Extractor}
In practice, the multimodal features of micro-videos are highly dimensional, redundant, and noisy, which cannot be fed directly to the meta-learning function. Towards this end, meta information extractor is employed to extract informative and concise meta information from multimodal features of the target item. In this work, we adopt a multi-layer perceptron (MLP) for the extraction,
\begin{equation}\label{eq:mlp}
\left\{
\begin{aligned}
    &\phi_{in}(\mathcal{X}_i)=\begin{bmatrix}\mathbf{x}_i^v\\\mathbf{x}_i^a\\\mathbf{x}_i^t\end{bmatrix}\\
    &\mathbf{s}_i = \phi_{out}\Big(\phi_{mid}\big(\phi_{in}(\mathcal{X}_i)\big)\Big)
\end{aligned}
\right.
,
\end{equation}
where $\phi_{in}$, $\phi_{mid}$, and $\phi_{out}$ respectively denote the input, hidden, and output layers of MLP, and each layer consists of its own weight matrix, bias vector, and activation function. To be more specific, we uniformly select ReLU as the activation function. After successive multi-layer processing, $\mathbf{s}_i\in\mathbb{R}^{d_s}$ indicates the vector of meta information, wherein the size $d_s$ is far less than the length of input feature $(d_v+d_a+d_t)$. The possibility of other extraction approaches that cannot be ruled out, such as the attention mechanism, will be further explored in future work.

\subsubsection{Meta Fusion Learner}
Meta fusion learner is responsible for rapidly parameterizing the multimodal function  based on the task-dependent meta information. More specifically, meta fusion learner learns the mapping from the meta information $\mathbf{s}_i$, derived from the multimodal features $\mathcal{X}_i$, to the weight matrix parameter $\mathbf{W}_i$. Mathematically, the most direct and practicable way of mapping a 1-D vector into a 2-D weight matrix is to multiply by a 3-D tensor. Hence, in this work, we partially implement meta fusion learner as:
\begin{equation}
    \mathbf{W}_i^*=\mathcal{T}\times_3\mathbf{s}_i,
\end{equation}
where $\mathcal{T}\in\mathbb{R}^{d_m\times (d_v+d_a+d_t)\times d_s}$ is the 3-D tensor parameter for meta fusion learner; $d_m$, $(d_v+d_a+d_t)$, and $d_s$ indicate the argument values of height, width, and depth of $\mathcal{T}$. The operator $\times_3$ denotes that the multiplication is being performed along the tensor's third dimension ($d_s$). The tensor depth is exactly set equal to the size of meta information vector $\mathbf{s}_i$, in order to make the tensor-vector multiplication along the direction of depth. Thus, the product $\mathbf{W}_i^*\in\mathbb{R}^{d_m\times (d_v+d_a+d_t)}$ is dynamically adaptive to fusing the multimodal features of the target item $i$.

Beyond this, we also reserve an item-shared weight matrix $\mathbf{W}\in\mathbb{R}^{d_m\times (d_v+d_a+d_t)}$, which is equivalent to the previous static fusion method. This parameter contributes to modeling the elementary and common fusion for all the micro-videos. In comparison, the item-specific weight matrix $\mathbf{W}_i^*$ can be regarded as learning or adaption for the multimodal fusion task of item $i$. By combining $\mathbf{W}$ and $\mathbf{W}_i^*$, we obtain the parameter underlying a one-layer neural network for addressing the multimodal fusion task of a specific item. The multimodal representation can be learned via such a dynamically parameterized network as:
\begin{equation}
\left\{
\begin{aligned}
    &\mathbf{W}_i = \mathbf{W}+\mathbf{W}_i^*\\
    &\mathbf{e}_i^m = \mathbf{W}_i\begin{bmatrix}\mathbf{x}_i^v\\\mathbf{x}_i^a\\\mathbf{x}_i^t\end{bmatrix}
    \end{aligned}
\right.
.
\end{equation}
From the viewpoint of meta-learning, the weight $\mathbf{W}$ denotes the internal parameter that is broadly suitable to many tasks, while the generated parameter $\mathbf{W}_i^*$ can be regarded as fine-tuning the layer weight parameter for quickly adapting to an individual task~\cite{finn2017model}.

\subsection{Deep Multi-Layer Fusion Framework}
Due to the multi-layer nature of deep neural networks, each successive layer is hypothesized to represent the information more abstractly. It is claimed that most functions that can be represented compactly by deep architectures cannot be represented by a compact shallow architecture~\cite{bengio2007scaling}. Hence, proposing a multi-layer version of the MetaMMF model is necessary. In brief, multi-layer MetaMMF jointly learns the weight matrix parameters for a succession of neural layers, which are used to fuse the multimodal features step by step. Please kindly note that we omit the bias parameters in a neural layer, without which the performance is still good. Similarly, all the parameters are dynamically generated by taking in the multimodal features of the target item,
\begin{equation}
    \{\mathbf{W}_i^n\}_{n=1}^N = \mathcal{G}(\mathcal{X}_i),
\end{equation}
where $N$ is the number of layers. Specifically, meta information $\mathbf{s}_i$ is likewise extracted from the concatenation of multimodal features as Eqn.(\ref{eq:mlp}). Subsequently, the weight matrices of all layers are determined as follows,
\begin{equation}
    \{\mathbf{W}_i^n\}_{n=1}^N = \{\mathbf{W}^n + \mathcal{T}^n\times_3\mathbf{s}_i\}_{n=1}^N,
\end{equation}
where the 3-D tensors $\{\mathcal{T}^n\}_{n=1}^N$ and the 2-D matrices $\{\mathbf{W}^n\}_{n=1}^N$ are all trainable parameters for learning the weight matrices $\{\mathbf{W}_i^n\}_{n=1}^N$ of $N$ layers, respectively. All the elements in the sequence of 3-D tensors $\{\mathcal{T}^n\}_{n=1}^N$ have consistent depths, which are equivalent to the length of the meta information vector $\mathbf{s}_i$. Obviously, multi-layer MetaMMF is an extension and stacking of the single-layer ones. In this way, each item can be provided with its individual multi-layer neural network for more complex multimodal fusion. The multimodal representation of item $i$ can be learned via its own $N$-layer neural network as,
\begin{equation}
    \mathbf{e}_i^m = \mathbf{W}_i^N\sigma\bigg(\mathbf{W}_i^{N-1}\sigma\Big(...\sigma\big(\mathbf{W}_i^2\sigma(\mathbf{W}_i^1\begin{bmatrix}\mathbf{x}_i^v\\\mathbf{x}_i^a\\\mathbf{x}_i^t\end{bmatrix})\big)...\Big)\bigg),
\end{equation}
where $\sigma$ denotes the activation function of the neural layer to implement the non-linearity of fusion, and we employ LeakyReLU in this work. In terms of the network structure design, we employ a tower pattern. The first hidden layer is the broadest, containing fewer neurons in each subsequent layer. As we progress through the layers, we reduce the number of neurons by half, ensuring that each layer has no less than the output layer size $d_m$. The rationale behind this design is that deeper layers with fewer hidden neurons can learn more abstract features from the data.

\subsection{Model Simplification}
\begin{figure*}[tbp]
\centering
\includegraphics[width=0.7\linewidth]{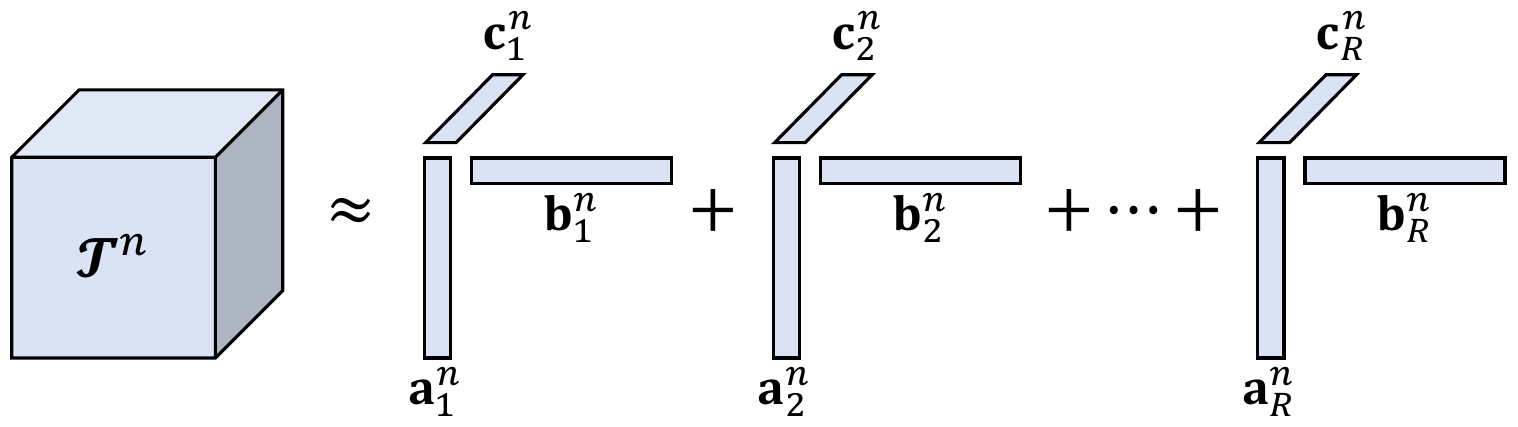}
\caption{CP decomposition of a 3-D tensor.}
\label{fig:tensor_cpd}
\end{figure*}

Although the multi-layer framework can model a more complex fusion function, the number of parameters inevitably increases as the layers stack. Compared with a static neural network, multi-layer MetaMMF is equipped with extra 3-D tensor parameters for dynamic parameterization, increasing the storage cost and training time. To address this issue, we propose a simplification method using canonical polyadic  (CP) decomposition to reduce the number of parameters in MetaMMF. Specifically, CP decomposition factorizes a tensor into a sum of component rank-one tensors, each of which is composed of the outer product of vectors. Through such decomposition, the dimension of the parameters can be greatly reduced. Therefore, we utilize CP decomposition to construct the 3-D tensor parameters used in MetaMMF. As illustrated in Figure~\ref{fig:tensor_cpd}, we can write the 3-D tensor $\mathcal{T}^n\in\mathbb{R}^{P\times Q\times Z}$ at the $n$-th layer as:
\begin{equation}\label{eq:cbd}
    \mathcal{T}^n\approx\sum_{r=1}^R\mathbf{a}_r^n\circ\mathbf{b}_r^n\circ\mathbf{c}_r^n,
\end{equation}
where notation $\circ$ denotes the outer product of vectors, $R$ is a positive integer and $\mathbf{a}_r^n\in\mathbb{R}^P$, $\mathbf{b}_r^n\in\mathbb{R}^Q$, and $\mathbf{c}_r^n\in\mathbb{R}^Z$ for $r=1, ..., R$. Elementwise, Eqn.(\ref{eq:cbd}) is written as,
\begin{equation}
    t_{pqz}^n\approx\sum_{r=1}^R a_{pr}^n b_{qr}^n c_{zr}^n, p=1, ..., P; q=1,...,Q; z=1,...,Z.
\end{equation}

It is proven that the tensor $\mathcal{T}^n$ can be approximatively decomposed into three factor matrices. The factor matrices refer to the combination of the vectors from the rank-one components, \textit{i.e.}, $\mathbf{A}^n=[\mathbf{a}_1^n~\mathbf{a}_2^n~ \cdots~\mathbf{a}_R^n]$ and likewise for $\mathbf{B}^n$ and $\mathbf{C}^n$. Using these definitions, we can replace the tensor $\mathcal{T}^n$ with the three matrices $\mathbf{A}^n\in\mathbb{R}^{P\times R}$, $\mathbf{B}^n\in\mathbb{R}^{Q\times R}$, and $\mathbf{C}^n\in\mathbb{R}^{Z\times R}$ to generate dynamic fusion parameters of the $n$-th layer as,
\begin{equation}
    \mathbf{W}_i^n=\mathbf{W}^n+\llbracket \mathbf{A}^n, \mathbf{B}^n, \mathbf{C}^n\rrbracket \times_3\mathbf{s}_i,
\end{equation}
where $\llbracket \cdot,\cdot,\cdot \rrbracket$ indicates the computing process in Eqn.(\ref{eq:cbd}). Comparing the space complexity before and after CP decomposition, we can find it decreasing from $\mathcal{O}(P\cdot Q\cdot Z)$ to $\mathcal{O}(R\cdot(P+Q+Z))$, since $Q>P\gg R$. When the number of tensors increases with fusion layers, the effect of model simplification will be more significant.

\subsection{Model-Agnostic Combination with MF \& GCN}
As described above, MetaMMF can dynamically provide a multimodal fusion function in the form of a neural network for each item, which integrates the features from multiple modalities into a joint multimodal representation. Benefiting from the end-to-end training of neural networks, MetaMMF can be seamlessly combined with other representation learning frameworks where the joint multimodal representations are used for initialization. In this way, the combined model can take advantage of both multimodal and other useful information, such as collaborative filtering (CF) information and graph structure information. To demonstrate the model-agnostic characteristic of MetaMMF, we present two combined methods to address the micro-video recommendation task.

\subsubsection{MetaMMF Plus MF}
Matrix factorization (MF) is one of the most simple and effective recommendation methods, only utilizing CF signals. MF learns two independent embedding matrices to represent users and items, whose multiplication is used to reconstruct the observed ratings and predict the undiscovered ratings. The joint multimodal representation learned by MetaMMF can be directly injected into the MF framework, and thus combining the multimodal information with the CF signal. Therefore, we propose a combined method named \textbf{MetaMMF\_MF}. Similar to VBPR~\cite{he2016vbpr}, we extend the item collaborative representation $\mathbf{e}_i^c$ with the joint multimodal representation $\mathbf{e}_i^m$ as the complete item representation $\mathbf{e}_i$, like Eqn.(\ref{eq:cat}). Following the MF framework, we estimate the affinity score between the target user and item by the inner product of their representations,
\begin{equation}\label{eq:y_ui}
    \hat{y}_{ui} = \mathbf{e}_u^\top\mathbf{e}_i = \mathbf{e}_u^\top\begin{bmatrix}\mathbf{e}_i^m\\\mathbf{e}_i^c\end{bmatrix},
\end{equation}
wherein $\mathbf{e}_u\in\mathbb{R}^{(d_m+d_c)}$ and $\mathbf{e}_i^c\in\mathbb{R}^{d_c}$ are both trainable embedding parameters corresponding to user $u$ and item $i$, responsible for capturing the CF signals.

\subsubsection{MetaMMF Plus GCN}
Graph Convolution Network (GCN) is a state-of-the-art representation learning technology that injects high-order connectivity signals into representations by recursively passing and aggregating messages based on the graph structure. In the recommendation task, GCN is always applied to the user-item interaction graph where the nodes correspond to users and items, connected by historical interactions~\cite{liu2021interest,liu2022hs}. Each node is initialized with an embedding, recursively updated by the convolution operation on its neighbor nodes. In most cases, the node initialization involves individual features, \textit{e.g.}, identity and content features. Hence, nodes can finally obtain higher-quality representations by aggregating information from multi-hop neighbors.

However, such graph convolution operation only conducts on unimodal representation. For information of multiple modalities, some methods construct multiple interaction graphs for all modalities, such as MMGCN~\cite{wei2019mmgcn}. In comparison, MetaMMF can integrate the information from multiple modalities into a joint representation, directly serving as the node initialization of an interaction graph. Therefore, we employ MetaMMF in the initial layer of GCN, and propose a combined method named \textbf{MetaMMF\_GCN}. Specifically, user $u$ and item $i$ can be initialized as:
\begin{equation}
    \mathbf{e}_u^{(0)}=\mathbf{e}_u,~\mathbf{e}_i^{(0)}=\begin{bmatrix}\mathbf{e}_i^m\\\mathbf{e}_i^c\end{bmatrix}
\end{equation}
where $\mathbf{e}_u^{(0)}$ and $\mathbf{e}_i^{(0)}$ denote the initial user and item representations at the $0$-th layer of GCN. Similarly, $\mathbf{e}_u^{(0)}$ is an entirely trainable embedding of user $u$, while $\mathbf{e}_i^{(0)}$ is obtained by concatenating the joint multimodal representation derived from MetaMMF and the trainable collaborative representation of item $i$. Based upon the representations at the $0$-th layer, we recursively formulate the item $i$'s representation at the $l$-th layer as:
\begin{equation}
    \mathbf{e}_i^{(l)}=\sigma(\mathbf{W}_1\mathbf{e}_i^{(l-1)}+ \mathbf{W}_2\cdot\text{mean}_{u\in\mathcal{N}_i}\mathbf{e}_u^{(l-1)}),
\end{equation}
where $\mathcal{N}_i$ denotes the neighborhood set of item $i$, \textit{i.e.}, the users whom item $i$ directly interacted with. $\sigma$ denotes the LeakyReLU activation function. Note that we utilize the most common graph convolution operation in~\cite{hamilton2017inductive}, and other more complicated choices are left to be explored. Analogously, we can obtain the representation $\mathbf{e}_u^{(l)}$ for user $u$ by propagating information from its connected items. For convenience, we denote the final representations at depth $L$ as $\mathbf{e}_u^{(L)}$ and $\mathbf{e}_i^{(L)}$. The inner product of the final user and item representations is used to predict their affinity score like most GCN-based methods do.

\subsection{Optimization}
In this work, we optimize the proposed models based on implicit feedback, such as clicks, views, and purchases. Compared to explicit ratings, implicit feedback is easier to collect but more challenging to model user preferences~\cite{guo2019attentive}, due to its scarcity of negative feedback. To learn model parameters, we optimize the pairwise BPR loss~\cite{rendle2009bpr}, which is often used in recommender systems. It considers the relative order between observed and unobserved user-item interactions. Specifically, BPR assumes that the items in observed interactions reflect a user’s preference more than those in unobserved ones and should be assigned with higher prediction scores. The objective function is as follows,
\begin{equation}
    \mathcal{L}=\sum_{(u,i,j)\in\mathcal{D}}-\text{ln}~\sigma(\hat{y}_{ui}-\hat{y}_{uj})+\lambda||\Phi||_2^2,
\end{equation}
where $\mathcal{D}=\{(u,i,j)|(u,i)\in\mathcal{R}^+,(u,j)\in\mathcal{R}^-\}$ denotes the pairwise training data, $\mathcal{R}^+$ indicates the observed interactions, and $\mathcal{R}^-$ is the unobserved interactions; $\sigma(\cdot)$ is the sigmoid function; $\Phi$ denotes all trainable model parameters, and $\lambda$ controls the $L_2$ regularization strength to prevent overfitting. We adopt mini-batch Adam~\cite{kingma2015adam} to optimize the prediction model and update the model parameters. In particular, for a batch of randomly sampled triples $(u,i,j)\in\mathcal{D}$, we establish their representations after MetaMMF and the subsequent MF or GCN, and then update model parameters by using the gradients of the loss function.
\section{Experimental Setup}
In this section, the evaluation datasets are presented first, followed by an explanation of the experimental settings and an elaboration of baseline methods.

\subsection{Datasets}
As micro-videos contain rich multimodal information --- frames, soundtracks, and descriptions, the experimental datasets should include such multimodal data for each item. Following MMGCN~\cite{wei2019mmgcn} and GRCN~\cite{wei2020graph}, we performed experiments on three publicly accessible datasets designed for micro-video personalized recommendation, including MovieLens\footnote{https://movielens.org/.}, TikTok\footnote{https://www.tiktok.com/.}, and Kwai\footnote{https://www.kwai.com/.}. Table~\ref{tab:dataset} summarizes the statistics of the datasets.

\begin{itemize}[leftmargin=*]
\item \textit{MovieLens:} This dataset is widely adopted to evaluate personalized recommendation. To adapt it for micro-video recommendation, we collected the descriptions of movies from the MovieLens-10M, and crawled the movies' trailers from Youtube\footnote{https://www.youtube.com/.}. Then, the pre-trained ResNet50~\cite{he2016deep} extracted the visual features from these trailers' keyframes. FFmpeg\footnote{http://ffmpeg.org/.} and VGGish~\cite{hershey2017cnn} were adopted to separate audio tracks and learn the acoustic features, respectively. Sentence2Vector~\cite{arora2017simple} was utilized to derive the textual features from movies' descriptions. Our experiments treated all ratings as the implicit feedback between the corresponding user-item pairs.
\item \textit{TikTok:} This dataset is released by the micro-video sharing platform Tiktok, which allows users to create and share micro-videos. It consists of users, micro-videos, and their interactions (\textit{e.g.}, clicks). The micro-video features in each modality are extracted and published without providing the raw data. Particularly, the textual features are extracted from the micro-video captions.
\item \textit{Kwai:} As a micro-video service provider, Kwai released a sizable micro-video dataset. The dataset contains users, micro-videos, and user behavior records with timestamps. To evaluate the proposed method from implicit feedback, we collected some click records of the corresponding users and micro-videos in a certain period. The acoustic features are unavailable, in contrast to the datasets mentioned above.
\end{itemize}

For each dataset, we randomly split the historical interactions of each user into three folds: 80\% for training, 10\% for validation, and the rest 10\% for testing. For each observed user-item interaction in the training set, we treated it as a positive instance. Then we adopted the negative sampling strategy to pair it with negative items the user did not interact with in the training set. This approach contrasts with that of~\cite{wei2020graph}, which used the entire dataset for negative sampling. To ensure a fair comparison, the negative sampling setup remains consistent across our methods and the compared baselines. The constructed triples are used for parameter optimization. The validation and testing sets are used to tune the hyper-parameters and evaluate the performance in the experiments, respectively.

\begin{table}[]
    \centering
    \caption{Statistics of the evaluation datasets. ($d_v$, $d_a$, and $d_t$ denote the dimensions of visual, acoustic, and textual modality feature data, respectively.)}
    \vspace{-0.35cm}
    \begin{tabular}{|c|c|c|c|c|c|c|c|}
         \hline
         Dataset & \#Users & \#Items & \#Interactions & Density & $d_v$ & $d_a$ & $d_t$ \\
         \hline
         \hline
         MovieLens & 55,485 & 5,986 & 1,239,508 & 0.37\% & 2048 & 128 & 100\\
         \hline
         TikTok & 36,656 & 76,085 & 726,065 & 0.03\% & 128 & 128 & 128\\
         \hline
         Kwai & 7,010 & 86,483 & 298,492 & 0.05\% & 2048 & - & 128\\
         \hline
    \end{tabular}
    \label{tab:dataset}
\end{table}

\subsection{Experimental Settings}
\textit{Evaluation Metrics:} For each user in the testing set, we regarded the items that the user did not interact with as the negative ones. Then each recommendation method predicts the user’s preference scores over all items, except the ones used in the training set. To measure the performance of top-$K$ recommendation and preference ranking, we ranked the items in descending order and adopted the widely-used evaluation metrics: Precision@$K$, Recall@$K$, HR@$K$ (Hit Ratio) and NDCG@$K$ (Normalized Discounted Cumulative Gain). 
By default, we set $K=10$ and $20$, respectively. We reported the average metrics of all users in the testing set.

\textit{Model Implementation:} 
We implemented the two combined models proposed above MetaMMF\_MF and MetaMMF\_GCN via the development tools Pytorch\footnote{https://pytorch.org/.} and Pytorch Geometric\footnote{https://pytorch-geometric.readthedocs.io/en/latest/.}. The user/item representation size is fixed at $64$. We optimized our model using the Adam optimizer with a batch size of 3,000, searching in batches of \{500, 1,000, 2,000, 3,000\}. Besides, we initialized all model parameters using the well-known Xavier approach. To tune the hyperparameters, we applied a grid search based on the results from the validation set. Specifically, we searched for the optimal learning rate from \{1e-4, 5e-4, 1e-3, 5e-3\} and ultimately set it to 1e-3. The coefficient $\lambda$ of $L_2$ normalization was searched within \{1e-8, 1e-7, · · ·, 1\}, and the optimal value was determined to be 1e-6. As another critical model hyperparameter, the length of the meta information vector, denoted by $d_s$, was fine-tuned among the values of \{2, 3, ..., 10\}, and eventually fixed at 5. Moreover, if Precision@$K$ on the validation set does not rise for ten successive epochs, early stopping is adopted. Experimentally, the optimal layer number for MetaMMF on the MovieLens dataset is $4$. The sequence of neuron sizes from the input layer to the output layer is as follows:  $2276\to 1024 \to 512 \to 256 \to 32$. On the TikTok and Kwai datasets, the optimal layer number for MetaMMF is $1$. The sequences of neuron sizes are $384 \to 32$ and $2176 \to 32$, respectively.

\subsection{Baseline Comparison}
To demonstrate the effectiveness of our method on micro-video recommendation, we compared it with the following state-of-the-art methods. These baselines can be briefly divided into three groups: MF-based (\textit{i.e.}, VBPR), GCN-based (\textit{i.e.}, GraphSAGE, NGCF, LightGCN, and GAT), and multimodal (MMGCN, GRCN, LATTICE, and InvRL) methods. 
\begin{itemize}[leftmargin=*]
\item \textbf{VBPR}~\cite{he2016vbpr}: This is a benchmark model in the multimodal recommendation. It incorporates content information with the MF framework to predict the interactions between users and items. For adapting to micro-video recommendation, we concatenated the multimodal features of micro-videos as content information.
\item \textbf{GraphSAGE}~\cite{hamilton2017inductive}: With the trainable aggregation function, this model can pass the message along the graph structure and collect them to update each node's representation. It considers both the structure information and the distribution of node features in the neighborhood. For making a fair comparison, we integrated multimodal features as node features to learn the representations.
\item \textbf{NGCF}~\cite{wang2019neural}: This model presents a novel recommendation framework to integrate the interaction information into the representation learning process. By exploiting the high-order connectivity from user-item interactions, the model encodes the CF signals into the representations. For a fair comparison, we fed the multimodal features of micro-videos into the framework to predict the user-item interactions.
\item \textbf{LightGCN}~\cite{he2020lightgcn}: This light yet effective model includes the essential ingredients of GCN for recommendation. This model adopts the simple weighted sum aggregator as a graph convolution operation and combines the representations obtained at each layer to form the final representation.
\item \textbf{GAT}~\cite{velivckovic2018graph}: This GCN-based method is able to automatically learn and specify different weights to the neighbors of each node. With the learned weights, it alleviates the noisy information from the neighbors to improve the GCN performance.
\item \textbf{MMGCN}~\cite{wei2019mmgcn}: This is a framework designed for micro-video recommendation. It allocates an independent GCN for each modality, which learns the modality-specific user preferences and item representations via the propagation of modality information on the user-item graph. The outputs of these GCNs are integrated as the final representations for recommendation.
\item \textbf{GRCN}~\cite{wei2020graph}: This is a state-of-the-art method for multimodal recommendation with implicit feedback. It adaptively adjusts the structure of the interaction graph according to the status of model training, instead of keeping the structure fixed. Based on the refined graph, it applies a graph convolution layer to distill informative signals on user preference.
\item \textbf{LATTICE}~\cite{zhang2021mining}: This is an advanced method for multimodal recommendation, which discovers the item relationship about multimodal features to learn a graph structure. Along the graph structure, graph convolutions can aggregate informative high-order affinities from neighbors for each item. Finally, the model makes recommendation by combining downstream CF methods.
\item \textbf{InvRL}~\cite{du2022invariant}: This is a state-of-the-art method for invariant representation learning in multimodal recommendation. InvRL can eliminate spurious correlations using heterogeneous environments to learn consistent invariant item representations across different environments. These invariant representations are then used to make accurate predictions of user-item interactions.
\end{itemize}

\textit{Baseline Implementation}. To ensure the consistency of baseline implementation, we adopted the publicly available implementations of   GraphSAGE\footnote{https://github.com/williamleif/GraphSAGE.}, NGCF\footnote{https://github.com/xiangwang1223/neural\_graph\_collaborative\_filtering.}, LightGCN\footnote{https://github.com/gusye1234/LightGCN-PyTorch.}, GAT\footnote{https://github.com/PetarV-/GAT.},  MMGCN\footnote{https://github.com/weiyinwei/MMGCN.}, GRCN\footnote{https://github.com/weiyinwei/GRCN.}, LATTICE\footnote{https://github.com/CRIPAC-DIG/LATTICE.}, and InvRL\footnote{https://github.com/nickwzk/InvRL.}. Without specification, the default size of user/item representation is $64$. For all the GCN-based baselines, we uniformly set a two-layer structure with parameters initialized from Xavier, and used the Adam optimizer with a well-chosen mini-batch size for model optimization. The learning rate is tuned amongst \{1e-4, 5e-4, 1e-3, 5e-3, 1e-2\}.

\section{Experimental Results}
To validate the effectiveness of our proposed method, we conducted extensive quantitative and qualitative experiments to answer the following research questions:
\begin{itemize}[leftmargin=*]
\item \textbf{RQ1}: Can our proposed method outperform the state-of-the-art baselines in the micro-video recommendation task?
\item \textbf{RQ2}: How does the depth of fusion layers affect MetaMMF?
\item \textbf{RQ3}: How do the representations benefit from the dynamic multimodal fusion?
\item \textbf{RQ4}: How do the CP decomposition of parameters affect the convergence and the performance of MetaMMF?
\item \textbf{RQ5}: How do the static weights affect MetaMMF?
\end{itemize}

\subsection{Performance Comparison (RQ1)}
\begin{table*}[htbp]\centering
\caption{Overall performance comparison between our model and the baselines on three datasets.}
\vspace{-0.3cm}
\setlength{\tabcolsep}{1.2mm}{
\begin{tabular}{|l|cc|cc|cc|cc|}
\hline
\multirow{2}*{Methods} & \multicolumn{8}{c|}{MovieLens} \\
\cline{2-9}
{} & P@10 & P@20 & R@10 & R@20 & HR@10 & HR@20 & N@10 & N@20 \\
\hline
\hline
VBPR & 0.0449 & 0.0348 & 0.1927 & 0.2875 & 0.1512 & 0.2340 & 0.1207 & 0.1494\\
\textbf{MetaMMF\_MF} & 0.0457 & 0.0353 &  0.1941 & 0.2912 & 0.1539 & 0.2375 & 0.1268 & 0.1562\\
\hline
GraphSAGE & 0.0493 & 0.0363 & 0.1979 & 0.3049 & 0.1601 & 0.2445 & 0.1351 & 0.1653\\
NGCF & 0.0539 & 0.0378 & 0.2132 & 0.3173 & 0.1658 & 0.2542 & 0.1366 & 0.1685\\
LightGCN & 0.0557 & 0.0391 & 0.2318 & 0.3359 & 0.1753 & 0.2646 & 0.1525 & 0.1845\\
GAT & 0.0564 & 0.0404 & 0.2250 & 0.3344 & 0.1813 & 0.2721 & 0.1523 & 0.1855\\
\hline
MMGCN & 0.0569 & 0.0420 & 0.2310 & 0.3522 & 0.1900 & 0.2851 & 0.1535 & 0.1977\\ 
GRCN & 0.0572 & 0.0424 & 0.2487 & 0.3558 & 0.1925 & 0.2867 & 0.1653 & 0.1999\\
LATTICE & \underline{0.0575} & \underline{0.0426} & 0.2500 & 0.3562 & \underline{0.1939} & \underline{0.2871} & \underline{0.1669} & \underline{0.2010}\\
InvRL & 0.0575 & 0.0424 & \underline{0.2518} & \underline{0.3584} & 0.1934 & 0.2869 & 0.1666 & 0.2008\\
\textbf{MetaMMF\_GCN} & \textbf{0.0599} & \textbf{0.0442} & \textbf{0.2613} & \textbf{0.3702} & \textbf{0.2018} & \textbf{0.2974} & \textbf{0.1757} & \textbf{0.2090}\\
\hline
\% Improv. & 4.17\% & 3.76\% & 3.77\% & 3.29\% & 4.07\% & 3.59\% & 5.27\% & 3.98\%\\
p-value & 1.12e-4 & 1.97e-4 & 1.85e-2 & 4.26e-3 & 2.38e-4 & 1.68e-3 & 3.05e-3 & 5.73e-3\\
\hline
\hline
\multirow{2}*{Methods} & \multicolumn{8}{c|}{TikTok} \\
\cline{2-9}
{} & P@10 & P@20 & R@10 & R@20 & HR@10 & HR@20 & N@10 & N@20 \\
\hline
\hline
VBPR & 0.0138 & 0.0111 & 0.0600 & 0.0945 & 0.0496 & 0.0799 & 0.0397 & 0.0500 \\
\textbf{MetaMMF\_MF} & 0.0143 & 0.0119 & 0.0673 & 0.1062 & 0.0521 & 0.0854 & 0.0422 & 0.0531\\
\hline
GraphSAGE & 0.0150 & 0.0124 & 0.0718 & 0.1134 & 0.0539 & 0.0893 & 0.0436 & 0.0560\\
NGCF & 0.0172 & 0.0141 & 0.0845 & 0.1335 & 0.0617 & 0.1014 & 0.0513 & 0.0658\\
LightGCN & 0.0184 & 0.0150 & 0.0921 & 0.1455 & 0.0705 & 0.1160 & 0.0565 & 0.0726\\
GAT & 0.0187 & 0.0152 & 0.0933 & 0.1474 & 0.0707 & 0.1163 & 0.0573 & 0.0736\\
\hline
MMGCN & 0.0186 & 0.0148 & 0.0927 & 0.1436 & 0.0706 & 0.1149 & 0.0570 & 0.0724\\ 
GRCN & 0.0195 & 0.0162 & 0.0948 & 0.1480 & 0.0717 & 0.1162 & 0.0586 & 0.0745\\
LATTICE & 0.0199 & 0.0164 & 0.0977 & 0.1535 & 0.0714 & 0.1177 & 0.0583 & 0.0754\\
InvRL & \underline{0.0203} & \underline{0.0166} & \underline{0.1002} & \underline{0.1559} & \underline{0.0726} & \underline{0.1179} & \underline{0.0612} & \underline{0.0779}\\
\textbf{MetaMMF\_GCN} & \textbf{0.0217} & \textbf{0.0172} & \textbf{0.1065} & \textbf{0.1612} & \textbf{0.0775} & \textbf{0.1237} & \textbf{0.0652} & \textbf{0.0816}\\
\hline
\% Improv. & 6.90\% & 3.61\% & 6.29\% & 3.40\% & 6.75\% & 4.92\% & 6.54\% & 4.75\%\\
p-value & 3.71e-3 & 7.81e-3 & 1.65e-3 & 2.08e-2 & 6.64e-4 & 3.92e-3 & 3.17e-3 & 6.95e-3\\
\hline
\hline
\multirow{2}*{Methods} & \multicolumn{8}{c|}{Kwai} \\
\cline{2-9}
{} & P@10 & P@20 & R@10 & R@20 & HR@10 & HR@20 & N@10 & N@20 \\
\hline
\hline
VBPR & 0.0108 & 0.0088 & 0.0302 & 0.0441 & 0.0220 & 0.0359 & 0.0221 & 0.0283\\
\textbf{MetaMMF\_MF} & 0.0139 & 0.0102 & 0.0388 & 0.0536 & 0.0285 & 0.0418 & 0.0336 & 0.0384\\
\hline
GraphSAGE & 0.0135 & 0.0105 & 0.0351 & 0.0490 & 0.0285 & 0.0429 & 0.0318 & 0.0364\\
NGCF & 0.0138 & 0.0106 & 0.0378 & 0.0553 & 0.0282 & 0.0434 & 0.0334 & 0.0391\\
LightGCN & 0.0153 & 0.0114 & 0.0417 & 0.0606 & 0.0314 & 0.0468 & 0.0357 & 0.0420\\
GAT & 0.0146 & 0.0111 & 0.0419 & 0.0609 & 0.0299 & 0.0455 & 0.0360 & 0.0417\\
\hline
MMGCN & 0.0140 & 0.0114& 0.0431 & 0.0614 & 0.0309 & 0.0466 & 0.0361 & 0.0419\\ 
GRCN & 0.0154 & 0.0116 & 0.0456 & 0.0649 & 0.0320 & 0.0479 & 0.0367 & 0.0428\\
LATTICE & 0.0156 & 0.0117 & 0.0460 & 0.0656 & 0.0323 & 0.0482 & 0.0379 & 0.0435\\
InvRL & \underline{0.0159} & \underline{0.0119} & \underline{0.0463} & \underline{0.0668} & \underline{0.0324} & \underline{0.0489} & \underline{0.0381} & \underline{0.0444}\\
\textbf{MetaMMF\_GCN} & \textbf{0.0166} & \textbf{0.0122} & \textbf{0.0480} & \textbf{0.0680} & \textbf{0.0340} & \textbf{0.0500} & \textbf{0.0398} & \textbf{0.0453}\\
\hline
\% Improv. & 4.40\% & 2.52\% & 3.67\% & 1.80\% & 4.94\% & 2.25\% & 4.46\% & 2.03\%\\
p-value & 3.44e-3 & 1.03e-2 & 4.50e-3 & 6.64e-3 & 1.05e-2 & 1.19e-2 & 4.51e-3 & 5.66e-3\\
\hline
\end{tabular}
} 
\label{tab:my_label}
\end{table*}

Table~\ref{tab:my_label} presents the results of our methods and baselines over the experimental datasets. Besides, it reports the improvements and statistical significance test, which are calculated between our proposed method and the strongest baseline (highlighted with underline). From the comparison results, we noted the following findings:
\begin{itemize}[leftmargin=*]
\item VBPR obtains poor performance on three datasets. This indicates that the static fusion function is insufficient to fit the distinct multimodal fusion of different micro-videos, and thus easily leads to suboptimal multimodal representations. Similarly under the MF framework, MetaMMF\_MF consistently outperforms VBPR across all cases, demonstrating the rationality and effectiveness of dynamic multimodal fusion. Despite obtaining rather competitive performance, MetaMMF\_MF is still slightly inferior to the GCN-based methods due to the limitation of MF.
\item Compared with the MF-based methods, it is obvious that representation learning can be further boosted via aggregation and propagation on the graph structure, since GCN-based methods show better performance on micro-video recommendation. More specifically, NGCF generally outperforms GraphSAGE since it injects high-order connectivities among users and items into its propagation layers. LightGCN achieves certain improvements over NGCF after removing some useless nonlinear operations in the original model. Meanwhile, GAT obtains remarkable performance attributed to the attention mechanism which can specify the adaptive weights of neighbor nodes.
\item As a model specifically designed for micro-video recommendation, MMGCN slightly outperforms the general GCN-based methods in most cases. One possible reason is that MMGCN sufficiently leverages the multimodal information by message passing of each modality and information interchange across modalities. Furthermore, GRCN adaptively adjusts the false-positive edges of the interaction graph to discharge the ability of GCN, thus outperforming MMGCN across all datasets. In addition to constructing the user-item graph for representation learning, LATTICE also creates modality-aware graphs that accurately capture the similarity of modal features between items, resulting in improved performance. Furthermore, the recently presented InvRL is confirmed to be the strongest baseline. InvRL is groundbreaking in that it learns invariant item representations to mitigate the influences of spurious correlations that are not adequately addressed by other methods.
\item MetaMMF\_GCN consistently yields the best performance across all datasets. In particular, the improvements of MetaMMF\_GCN over the strongest competitor \textit{w.r.t.} NDCG@$10$ are 5.27\%, 6.54\%, and 4.46\% in MovieLens, TikTok, and Kwai, respectively. Additionally, we conducted one-sample t-tests, which reveal that the improvements of MetaMMF\_GCN over the strongest baseline are statistically significant (p-value $<0.05$). The primary difference between MetaMMF and baselines lies in whether the multimodal fusion is dynamic or static. Hence, we attribute the significant improvements to our dynamic fusion strategy. MetaMMF can effectively model the complex relationships among the modalities in different items during representation learning by dynamically learning the item-specific fusion networks. This underscores the importance of dynamic multimodal fusion in micro-video recommendation.
\item Moreover, MetaMMF\_GCN significantly outperforms its MF version. It is reasonable since MetaMMF\_GCN further boosts the multimodal representations learned from MetaMMF via the message passing function of GCN, while MetaMMF\_MF directly uses the raw representations. This demonstrates that MetaMMF can be combined with other representation learning frameworks, and achieve better performance by exploiting the advantages of both frameworks to the full. 
\end{itemize}

\subsection{Effect of Layer Number on MetaMMF (RQ2)}
\begin{table*}[t]\centering
\caption{Effect of fusion neural layer numbers.}
\vspace{-0.3cm}
\begin{tabular}{|l|cc|cc|cc|}
\hline
\multirow{2}*{Methods} & \multicolumn{2}{c|}{MovieLens} & \multicolumn{2}{c|}{TikTok} & \multicolumn{2}{c|}{Kwai}\\
\rule{0pt}{10pt}
{} & P@10 & R@10 & P@10 & R@10 & P@10 & R@10 \\
\hline
\hline
Early\_MF & \underline{0.0448} & \underline{0.1923} & \underline{0.0137} & 0.0639 & \underline{0.0118} & \underline{0.0326} \\
Late\_MF & 0.0399 & 0.1730 & 0.0131 & 0.0598 & 0.0108 & 0.0304\\
Hybrid\_MF & 0.0407 & 0.1766 & 0.0134 & \underline{0.0651} & 0.0117 & 0.0321 \\
\hline
MetaMMF\_MF-1 & 0.0443 & 0.1844 & \textbf{0.0143} & \textbf{0.0673} & \textbf{0.0139} & \textbf{0.0388} \\
MetaMMF\_MF-2 & 0.0447 & 0.1884 & 0.0134 & 0.0609 & 0.0122 & 0.0353 \\
MetaMMF\_MF-3 & 0.0455 & 0.1930 & 0.0128 & 0.0580 & 0.0119 & 0.0349 \\
MetaMMF\_MF-4 & \textbf{0.0457} & \textbf{0.1941} & 0.0124 & 0.0584 & 0.0110 & 0.0337 \\
\hline
\% Improv. & 2.01\% & 0.94\% & 4.38\% & 3.38\% & 17.80\% & 19.02\%\\
p-value & 3.37e-4 & 1.02e-4 & 3.13e-4 & 2.38e-4 & 6.74e-3 & 3.54e-3\\
\hline
\hline
Early\_GCN & \underline{0.0573} & 0.2446 & 0.0186 & \underline{0.0925} & \underline{0.0151} & \underline{0.0406} \\
Late\_GCN & 0.0568 & 0.2264 & 0.0188 & 0.0890 & 0.0140 & 0.0386 \\
Hybrid\_GCN & 0.0572 & \underline{0.2477} & \underline{0.0189} & 0.0884 & 0.0151 & 0.0396 \\
\hline
MetaMMF\_GCN-1 & 0.0584 & 0.2554 & \textbf{0.0217} & \textbf{0.1065} & \textbf{0.0166} & \textbf{0.0480} \\ 
MetaMMF\_GCN-2 & 0.0592 & 0.2587 & 0.0214 & 0.1047 & 0.0160 & 0.0452 \\
MetaMMF\_GCN-3 & 0.0595 & 0.2590 & 0.0211 & 0.1043 & 0.0153 & 0.0439 \\
MetaMMF\_GCN-4 & \textbf{0.0599} & \textbf{0.2613} & 0.0211 & 0.1040 & 0.0151 & 0.0438 \\
\hline
\% Improv. & 4.54\% & 5.49\% & 14.81\% & 15.14\% & 9.93\% & 18.23\%\\
p-value & 3.03e-4 & 1.97e-3 & 2.01e-3 & 1.66e-3 & 2.62e-3 & 3.99e-3\\
\hline
\hline
\multirow{2}*{Methods} & \multicolumn{2}{c|}{MovieLens} & \multicolumn{2}{c|}{TikTok} & \multicolumn{2}{c|}{Kwai}\\
\rule{0pt}{10pt}
{} & HR@10 & N@10 & HR@10 & N@10 & HR@10 & N@10\\
\hline
\hline
Early\_MF & \underline{0.1508} & \underline{0.1204} & \underline{0.0495} & 0.0392 & \underline{0.0242} & 0.0282\\
Late\_MF & 0.1341 & 0.1120 & 0.0471 & 0.0391 & 0.0219 & 0.0225\\
Hybrid\_MF & 0.1370 & 0.1141 & 0.0482 & \underline{0.0401} & 0.0239 & \underline{0.0286}\\
\hline
MetaMMF\_MF-1 & 0.1491 & 0.1208 & \textbf{0.0521} & \textbf{0.0422} & \textbf{0.0285} & \textbf{0.0336}\\
MetaMMF\_MF-2 & 0.1516 & 0.1228 & 0.0515 & 0.0414 & 0.0249 & 0.0318\\
MetaMMF\_MF-3 & 0.1534 & 0.1252 & 0.0482 & 0.0379 & 0.0243 & 0.0283\\
MetaMMF\_MF-4 & \textbf{0.1539} & \textbf{0.1268} & 0.0462 & 0.0307 & 0.0235 & 0.0272\\
\hline
\% Improv. & 2.06\% & 5.32\% & 5.25\% & 5.24\% & 17.77\% & 17.48\%\\
p-value & 4.90e-4 & 6.09e-4 & 6.76e-4 & 5.47e-4 & 1.23e-2 & 7.05e-3\\
\hline
\hline
Early\_GCN & \underline{0.1929} & 0.1654 & \underline{0.0704} & \underline{0.0568} & \underline{0.0310} & 0.0346\\
Late\_GCN & 0.1837 & 0.1595 & 0.0675 & 0.0539 & 0.0287 & 0.0324\\
Hybrid\_GCN & 0.1925 & \underline{0.1671} & 0.0678 & 0.0543 & 0.0309 & \underline{0.0348} \\
\hline
MetaMMF\_GCN-1 & 0.1965 & 0.1709 & \textbf{0.0775} & \textbf{0.0652} & \textbf{0.0340} & \textbf{0.0398}\\ 
MetaMMF\_GCN-2 & 0.1992 & 0.1731 & 0.0767 & 0.0644 & 0.0327 & 0.0380\\
MetaMMF\_GCN-3 & 0.2003 & 0.1736 & 0.0756 & 0.0640 & 0.0314 & 0.0375\\
MetaMMF\_GCN-4 & \textbf{0.2018} & \textbf{0.1757} & 0.0757 & 0.0630 & 0.0308 & 0.0374\\
\hline
\% Improv. & 4.61\% & 5.15\% & 10.09\% & 14.79\% & 9.68\% & 14.37\%\\
p-value & 7.52e-4 & 2.59e-3 & 4.58e-3 & 5.64e-3 & 1.87e-3 & 5.43e-3\\
\hline
\end{tabular}
\label{tab:layer}
\vspace{-0.3cm}
\end{table*}

To verify the fusion advantage of MetaMMF, we compared it with neural network-based early fusion~\cite{mroueh2015deep}, late fusion~\cite{huang2013learning}, and hybrid fusion~\cite{lan2014multimedia} methods. We connected the output of these fusion methods with MF and GCN, respectively. Additionally, we examined the impact of layer numbers on MetaMMF's performance, considering the crucial role played by the meta-learned neural layer in MetaMMF. In particular, we varied the model depth in the $\{1, 2, 3, 4\}$ range. The experimental results of these methods are summarized in Table~\ref{tab:layer}, where MetaMMF\_MF-3 and MetaMMF\_GCN-3 denote our two methods with three meta-learned fusion layers, and similar notations are used for the others. By analyzing the results presented in Table~\ref{tab:layer}, we have made the following findings:
\begin{itemize}[leftmargin=*]
\item On the datasets with less sparsity (\textit{e.g.}, MovieLens), increasing the depth of MetaMMF substantially improves the performance when combined with both MF and GCN. The phenomenon validates that multi-layer MetaMMF is capable of modeling fine-grained multimodal fusion for items and thus enhances representation learning. On the sparse datasets (\textit{e.g.}, TikTok and Kwai), increasing the number of meta-learned fusion layers has no positive effect on the performance of MetaMMF. Especially when combined with MF, one-layer fusion achieves the best performance, which is consistent with the prior VBPR~\cite{he2016vbpr}. The probable reason is that MF easily suffers from over-fitting with sparse data, and redundant fusion layers actually make this problem worse. When followed by GCN, stacking fusion layers also has little impact on model performance. This is because the message passing of GCN can effectively alleviate the data sparsity problem with just one fusion layer~\cite{wang2019neural}.
\item Among the three kinds of fusion methods, early fusion consistently outperforms late fusion. Early fusion is the feature-level fusion that directly integrates multimodal features to learn the representation. By contrast, late fusion is at the decision level, overlooking the feature-level fusion among the modalities. This shows that late fusion is inapplicable to micro-video recommendation based on representation learning. Although the hybrid fusion method exploits both fusion methods in a common framework, it is inferior to the single early fusion in some cases. The possible reason is that the late fusion part drags down the performance.
\item We performed one-sample t-tests on various datasets and metrics to compare the best fusion strategies (highlighted with underlines) with our methods that employ optimal layers (highlighted in bold). The results are recorded in Table~\ref{tab:layer}, where a p-value < 0.05 indicates a statistically significant improvement. In the MF framework, we observed that MetaMMF significantly outperforms the early, late, and hybrid fusion methods when utilizing the optimal number of fusion layers. On the other hand, in the GCN framework, MetaMMF consistently outperforms all other fusion strategies, regardless of the number of fusion layers employed. This suggests that MetaMMF is better suited for collaborating with complex models, such as GCN, than other static fusion strategies. Our findings confirm the effectiveness of MetaMMF, demonstrating that dynamic multimodal fusion can significantly enhance the micro-video recommendation task.
\end{itemize}

\begin{figure}[t]
\centering
\begin{minipage}[t]{0.33\linewidth}
\centering
\includegraphics[width=\linewidth]{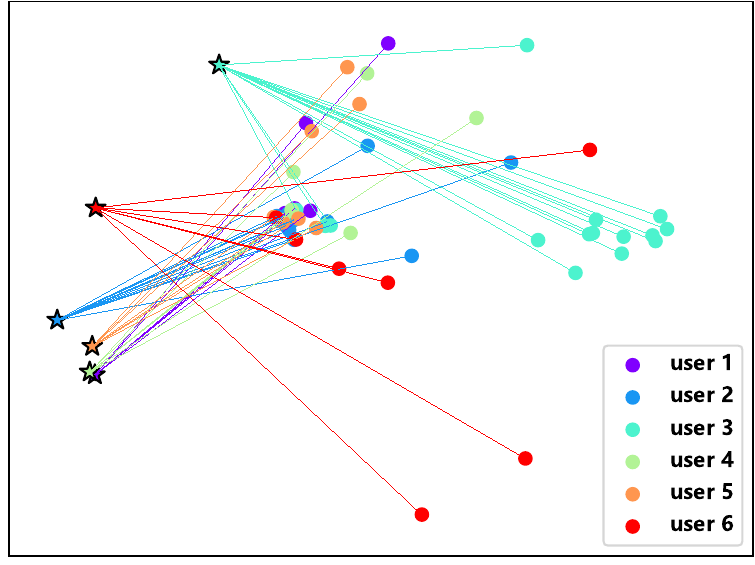}
\centerline{(a)~VBPR~($S = -0.0624$)}
\end{minipage}
\hspace{1cm}
\begin{minipage}[t]{0.33\linewidth}
\centering
\includegraphics[width=\linewidth]{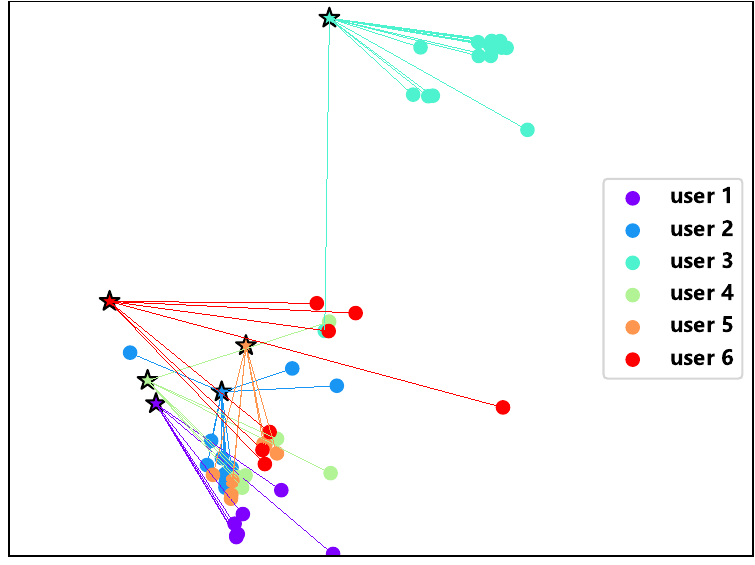}
\centerline{(b)~MetaMMF\_MF~($S = 0.0272$)}
\end{minipage}
\\
\vspace{0.5cm}
\begin{minipage}[t]{0.33\linewidth}
\centering
\includegraphics[width=\linewidth]{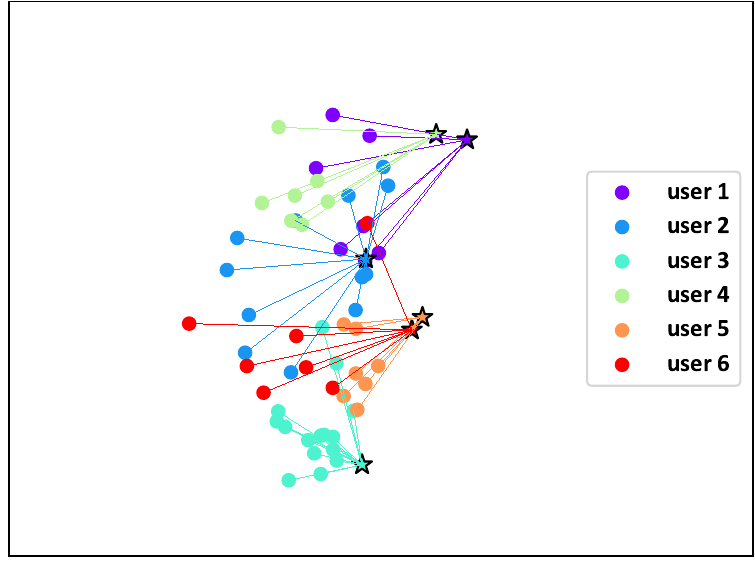}
\centerline{(c)~InvRL~($S = 0.0753$)}
\end{minipage}
\hspace{1cm}
\begin{minipage}[t]{0.33\linewidth}
\centering
\includegraphics[width=\linewidth]{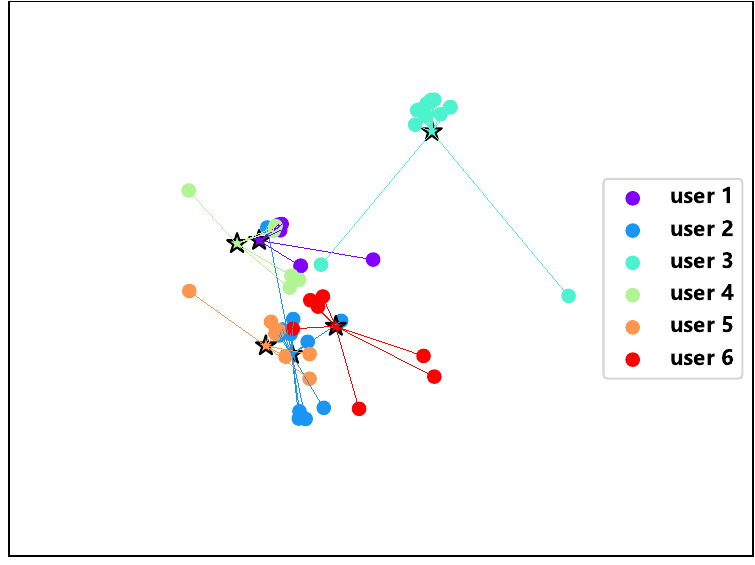}
\centerline{(d)~MetaMMF\_GCN~($S = 0.1248$)}
\end{minipage}
\centering
\vspace{-0.2cm}
\caption{The visualization depicts the t-SNE transformed representations obtained from our methods and baselines. Each star corresponds to a user from the TikTok dataset, while points with the same color signify relevant items. A link between a star and a point represents their interaction. For optimal viewing, please refer to the colored version. The notation $S$ indicates the mean silhouette coefficient of the clustering result for the sample representations.}
\label{fig:visual}
\vspace{-0.3cm}
\end{figure}

\subsection{Effect of Dynamic Multimodal Fusion (RQ3)}
In this section, we attempt to understand how our dynamic multimodal fusion strategy facilitates representation learning in the embedding space. Therefore, we randomly selected six users from TikTok associated with their relevant items. Note that the items are from the test set, which are not paired with users in the training phase. By employing the t-Distributed Stochastic Neighbor Embedding (t-SNE)~\cite{van2008visualizing} in 2-dimension, Figures~\ref{fig:visual}(a)-(d) visualize the representations derived from VBPR, MetaMMF\_MF, InvRL, and MetaMMF\_GCN, respectively. By analyzing how their representations distribute, we have two key observations:

\begin{itemize}[leftmargin=*]
\item The connectivities of users and items are well reflected in the embedding space, \textit{i.e.}, embedded into the nearest parts of the space, resulting in discernible clustering. Compared to VBPR and InvRL, our MetaMMF\_MF and MetaMMF\_GCN models exhibit even more noticeable clustering, with points of the same color (\textit{i.e.}, items interacted with by the same users) forming clusters. To evaluate the clustering performance of our methods and baselines, we computed their mean silhouette coefficients ($S$) for all samples shown in Figure~\ref{fig:visual}. This coefficient measures the effectiveness and reasonableness of the clustering, with a value ranging between $-1$ and $1$. A larger value indicates that samples from the same class are closer, while samples from different classes are farther apart, implying better clustering performance. Our methods' representations achieve better clustering performance than the competitors, as seen in the $S$ values in Figure~\ref{fig:visual}. This indicates that our MetaMMF model enhances representation learning effectively. We attribute this success to our model's dynamic multimodal fusion, which utilizes and integrates items' multimodal features in an adaptive approach.
\item Jointly analyzing the same users across Figure~\ref{fig:visual}, we find that the embeddings of their historical items tend to be closer in our methods. Meanwhile, the dynamic multimodal fusion prompts the item embeddings to distribute more discriminately in the visualization. It verifies that our proposed methods have an advantage in capturing the user-item correlation.
\end{itemize}

\subsection{Effect of Parameter Decomposition (RQ4)}
We investigate how the parameter decomposition (\textit{i.e.}, CP decomposition of 3-D tensor parameters) affects model MetaMMF from two aspects of performance and convergence. In particular, we first study how the hyper-parameter $R$ (\textit{i.e.}, the number of components in Eqn.(\ref{eq:cbd})) from CP decomposition affects the performance, and then explore the influence of CP decomposition on training epochs. 

\begin{figure}[tbp]
\centering
\begin{minipage}[t]{0.34\linewidth}
\centering
\includegraphics[width=\linewidth]{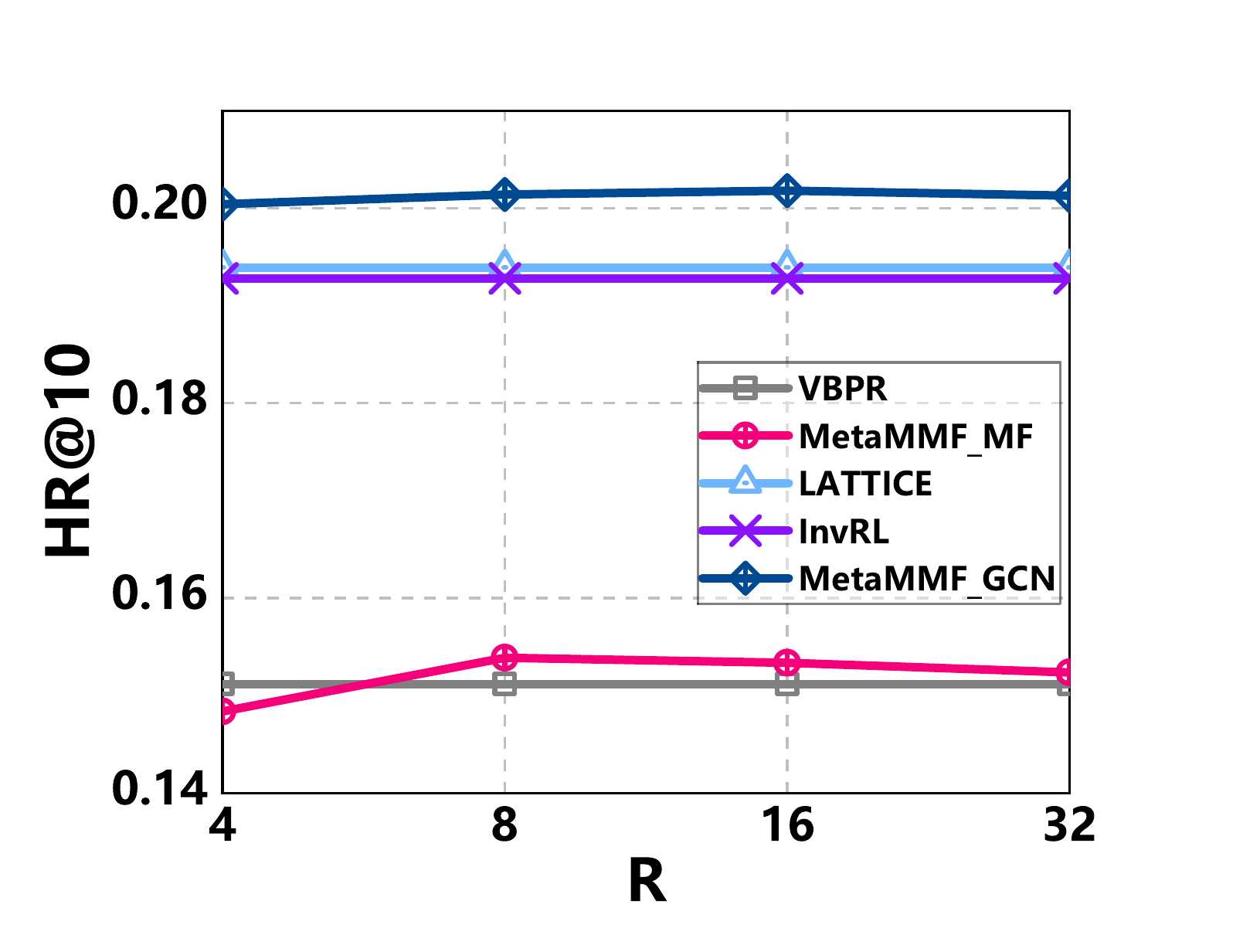}
\end{minipage}
\hspace{1cm}
\begin{minipage}[t]{0.34\linewidth}
\centering
\includegraphics[width=\linewidth]{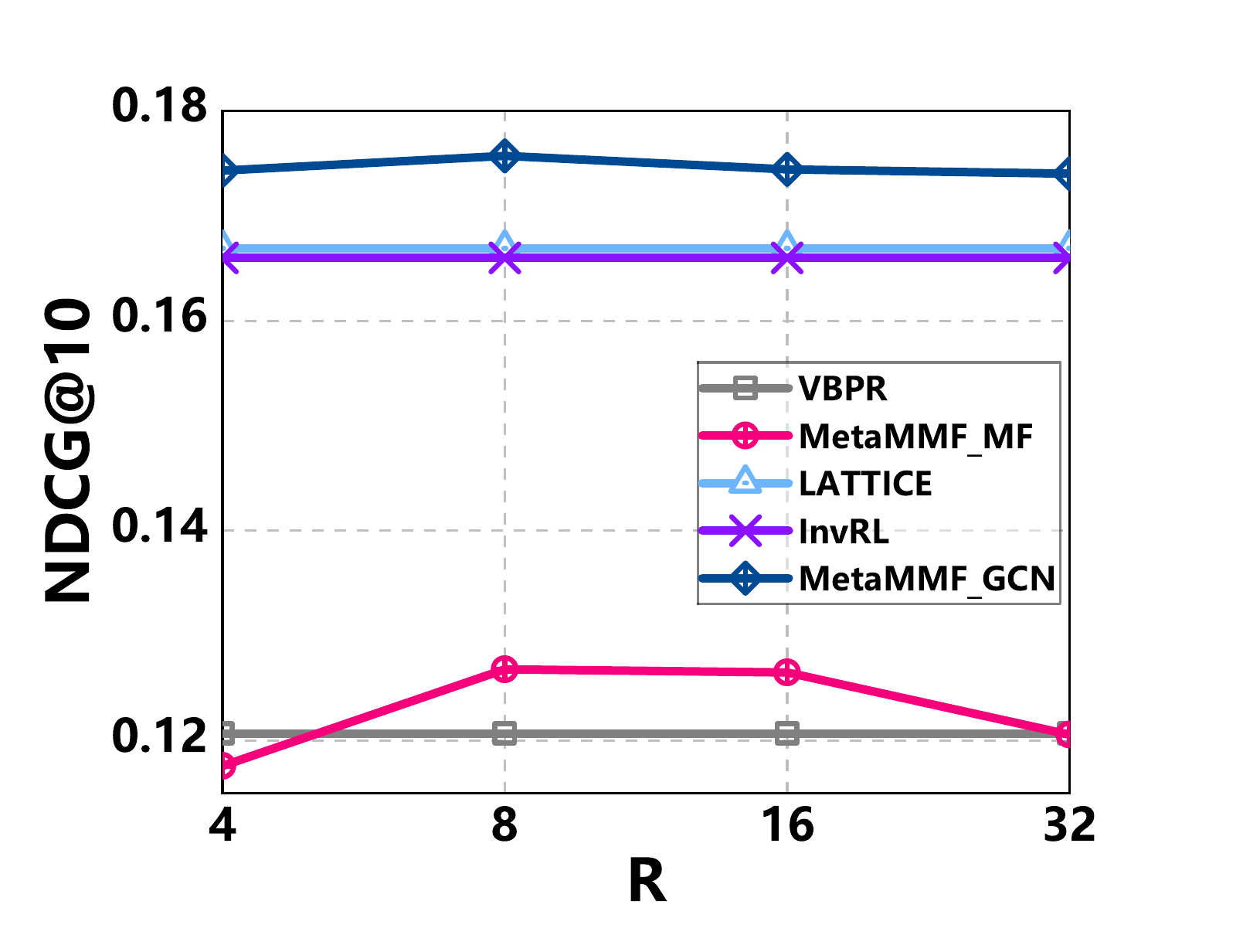}
\end{minipage}
\centerline{MovieLens}
\\
\begin{minipage}[t]{0.34\linewidth}
\centering
\includegraphics[width=\linewidth]{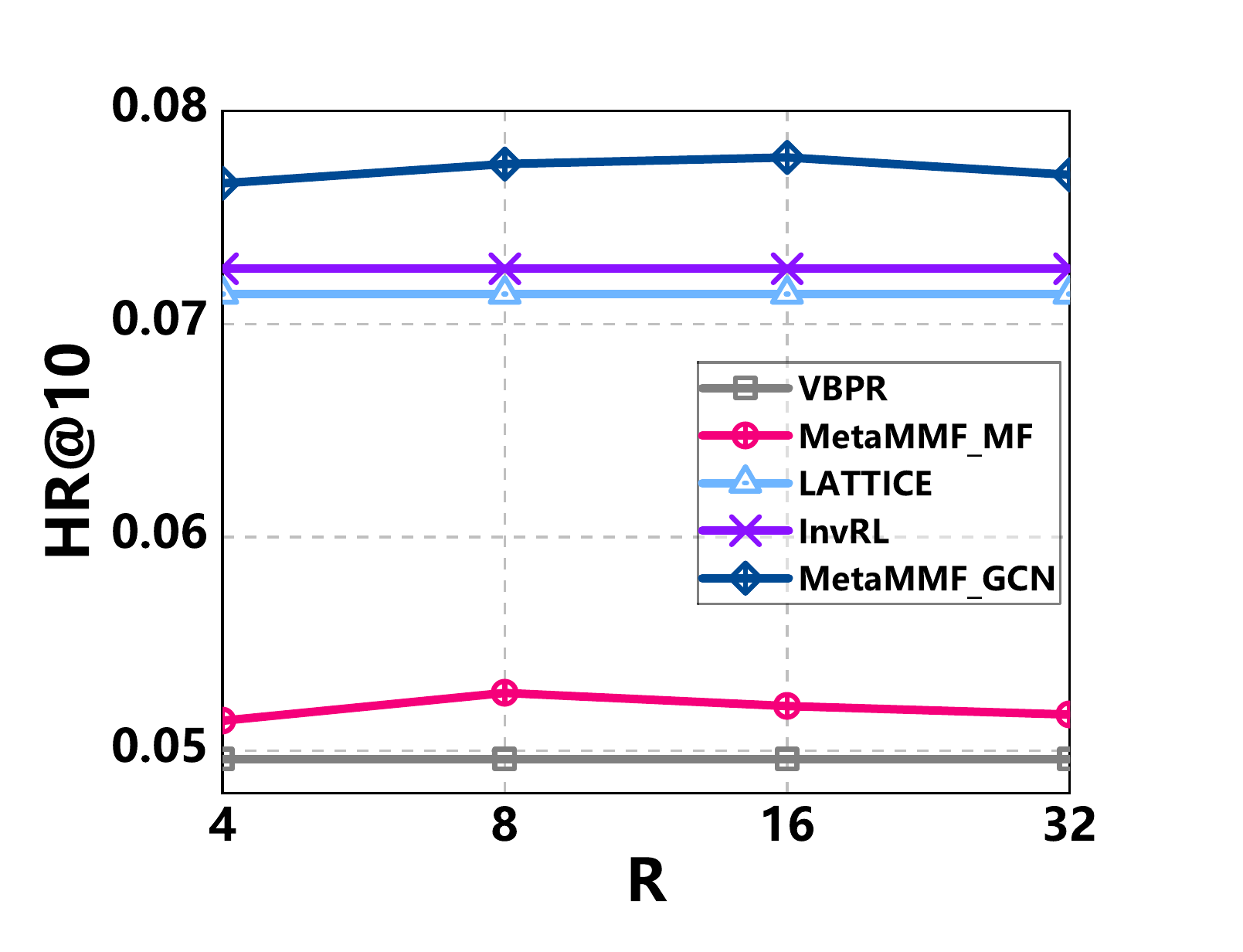}
\end{minipage}
\hspace{1cm}
\begin{minipage}[t]{0.34\linewidth}
\centering
\includegraphics[width=\linewidth]{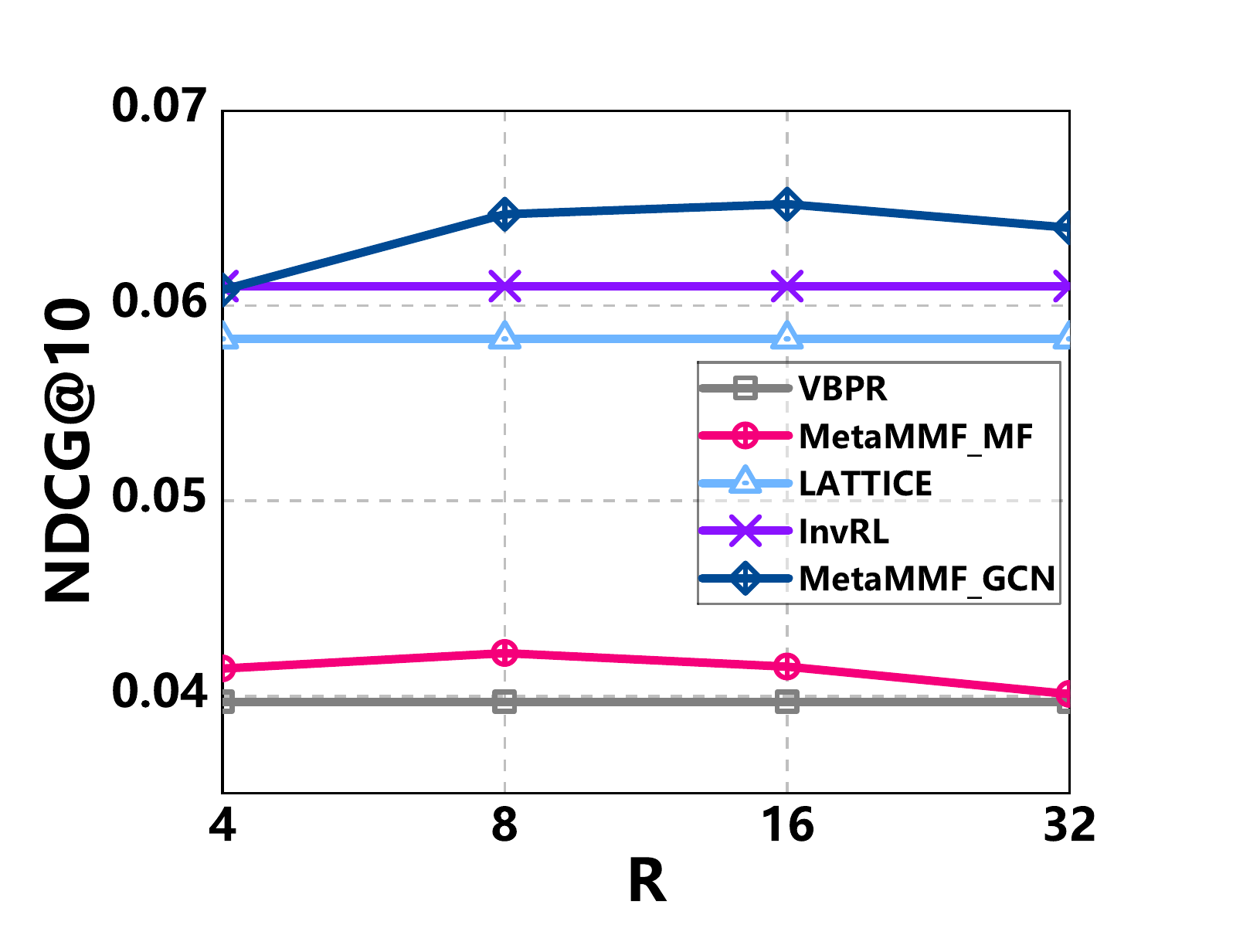}
\end{minipage}
\centerline{TikTok}
\\
\begin{minipage}[t]{0.34\linewidth}
\centering
\includegraphics[width=\linewidth]{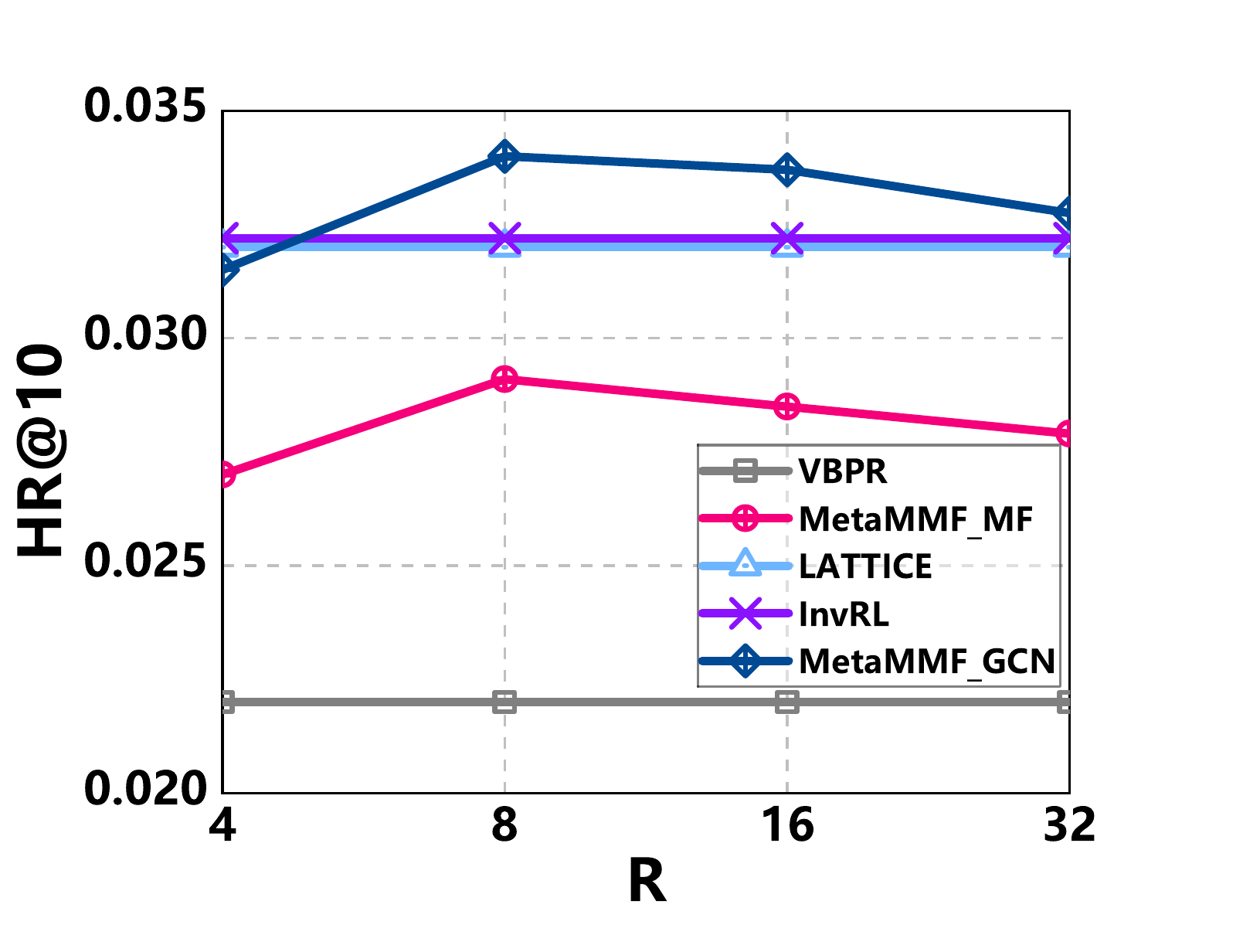}
\end{minipage}
\hspace{1cm}
\begin{minipage}[t]{0.34\linewidth}
\centering
\includegraphics[width=\linewidth]{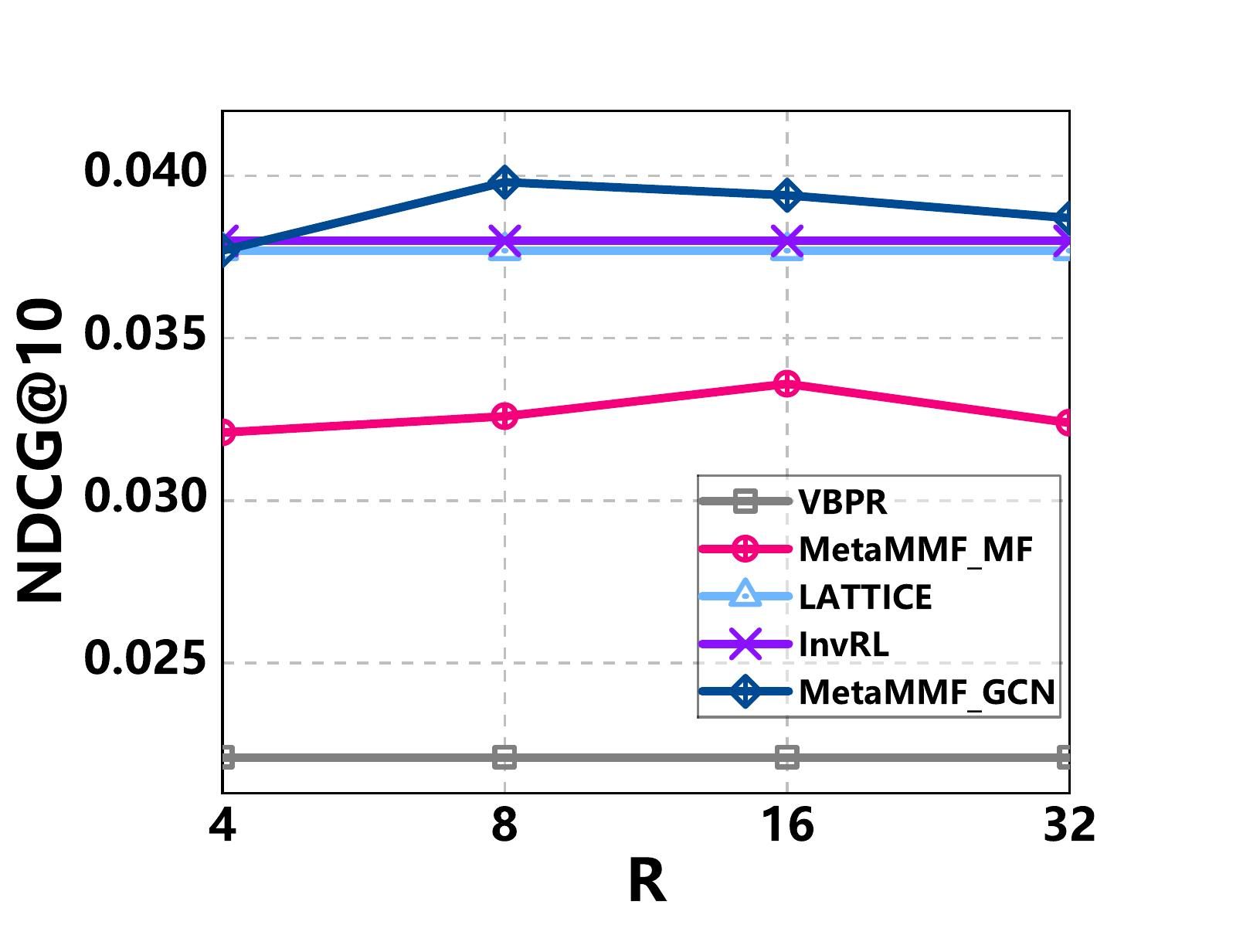}
\end{minipage}
\centerline{Kwai}
\centering
\vspace{-0.4cm}
\caption{Effect of hyper-parameter $R$ in CP decomposition.}
\label{fig:cpd}
\vspace{-0.5cm}
\end{figure}

\subsubsection{Effect of hyper-parameter $R$}
Figure~\ref{fig:cpd} shows the performance of our methods MetaMMF\_MF and MetaMMF\_GCN with different numbers of decomposed components, as well as their most competitive baselines, against HR@$10$ and NDCG@$10$ evaluation protocols on different datasets. The value of hyper-parameter $R$ is varied in the range of $\{4, 8, 16, 32\}$, while the number of fusion layers is optimal. From Figure~\ref{fig:cpd}, we can observe that:
\begin{itemize}[leftmargin=*]
\item The hyper-parameter $R$ in CP decomposition indeed affects the performance of our methods. Generally, a small decomposition value leads to the worst performance across all the datasets, since such large-scale decomposition cannot restore the fitting ability of the original 3-D tensor parameters. By increasing the value of $R$, the CP decomposition can offer better performance while effectively reducing our model's space complexity. However, blindly increasing the $R$ value will not further enhance the performance, but instead burdening the model. The possible reason is that $R$ has already been higher than the rank of the original 3-D tensor, and CP decomposition provides excess capacity.
\item Despite the influence of $R$, our methods are consistently superior to VBPR and LATTICE, respectively. This demonstrates that CP decomposition can replace the original 3-D tensor parameters, and retain the ability of dynamic multimodal fusion within our model. With regard to the optimal value of $R$, we find that MetaMMF\_MF and MetaMMF\_GCN always achieve the best performance on datasets when the value of $R$ is set to $8$ or $16$. In such a case, CP decomposition is able to reduce the model to nearly one percent of its original size. Hence, although MetaMMF uses 3-D tensor parameters to replace the 2-D matrix parameters in neural layers, the number of model parameters does not exceed by a wide margin attributing to CP decomposition.
\end{itemize}

\subsubsection{Effect of CP Decomposition on Model Convergence}
\begin{figure}[tbp]
\centering
\begin{minipage}[t]{0.33\linewidth}
\centering
\includegraphics[width=\linewidth]{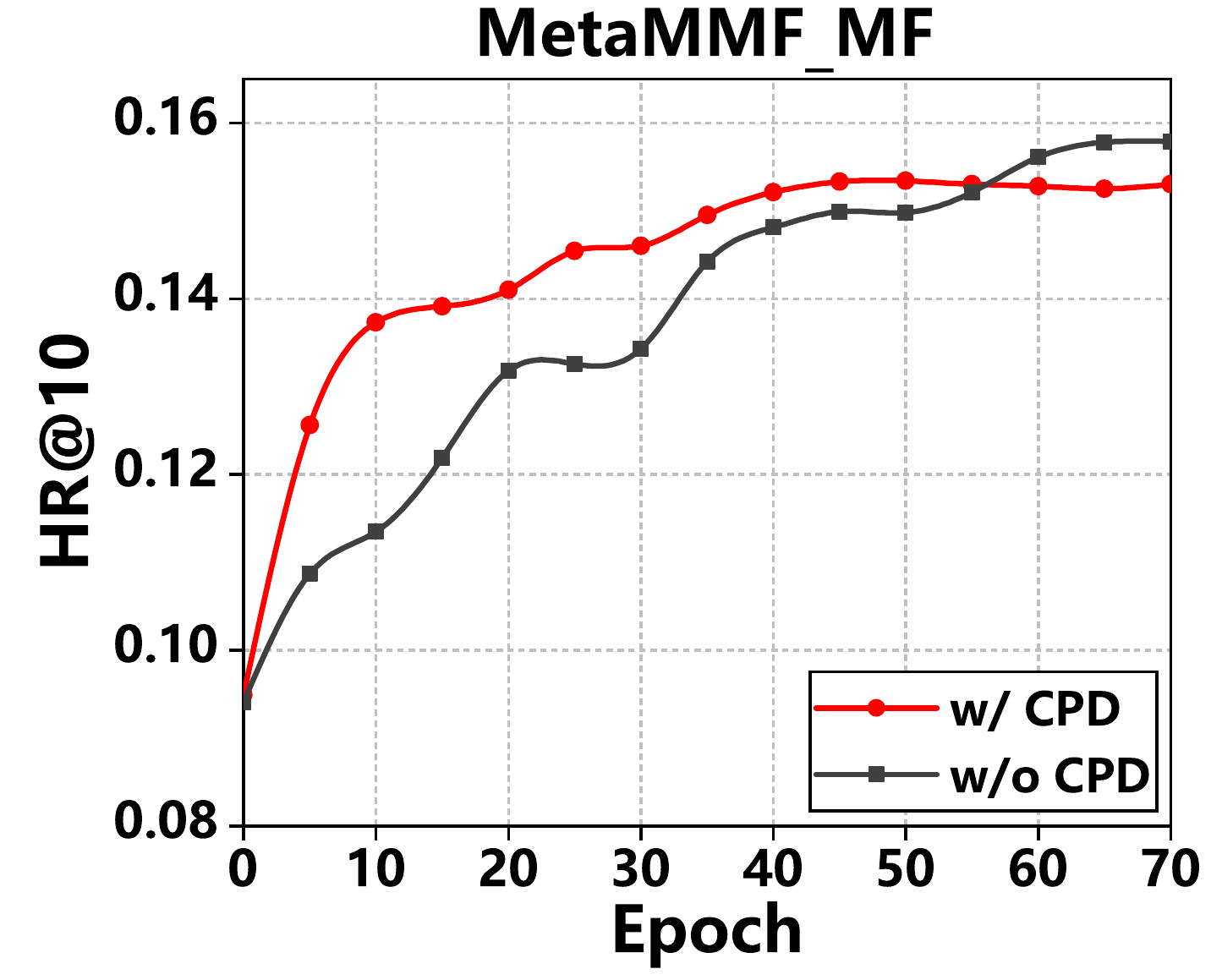}
\end{minipage}
\hspace{1cm}
\begin{minipage}[t]{0.33\linewidth}
\centering
\includegraphics[width=\linewidth]{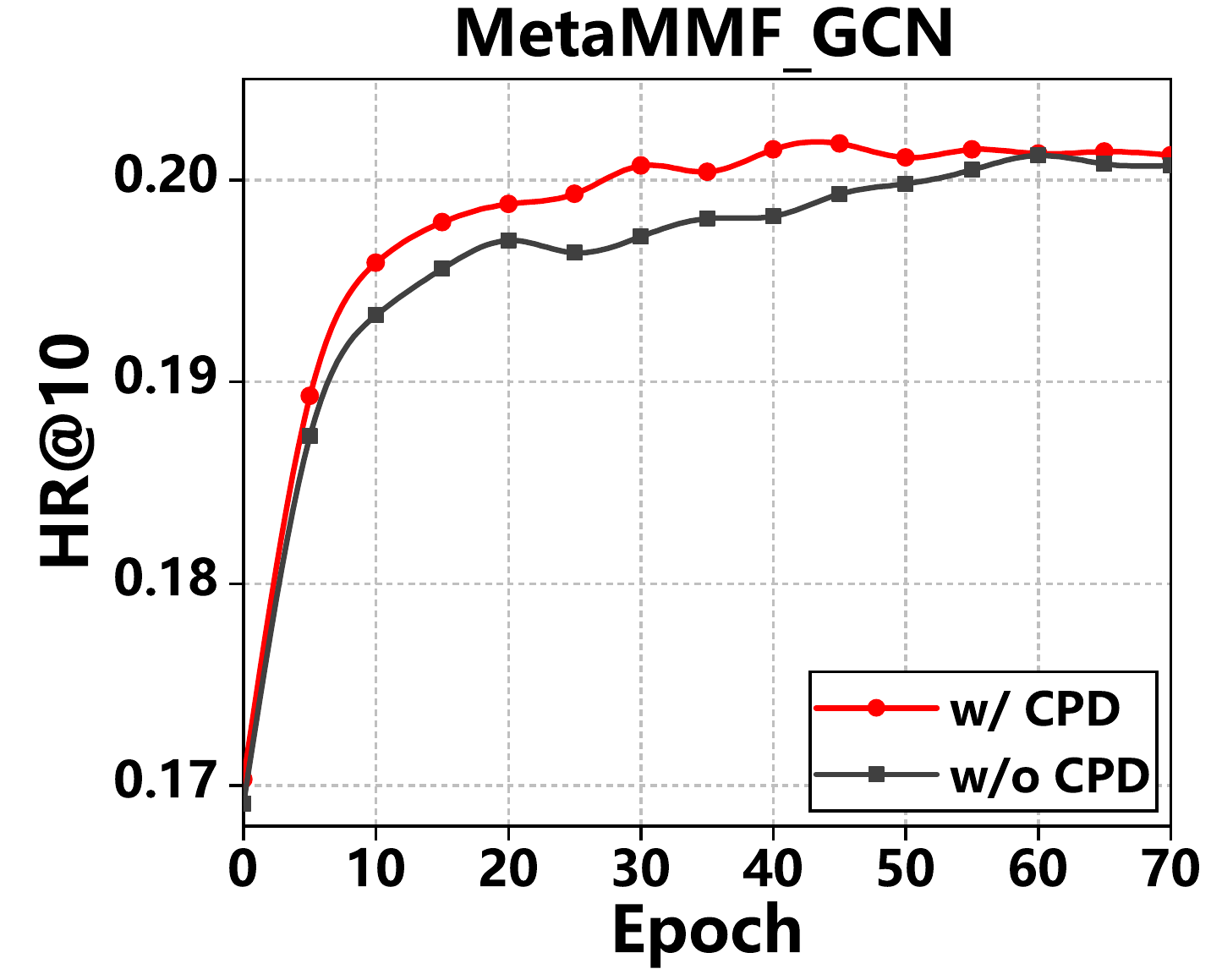}
\end{minipage}
\centerline{MovieLens}
\\
\begin{minipage}[t]{0.33\linewidth}
\centering
\includegraphics[width=\linewidth]{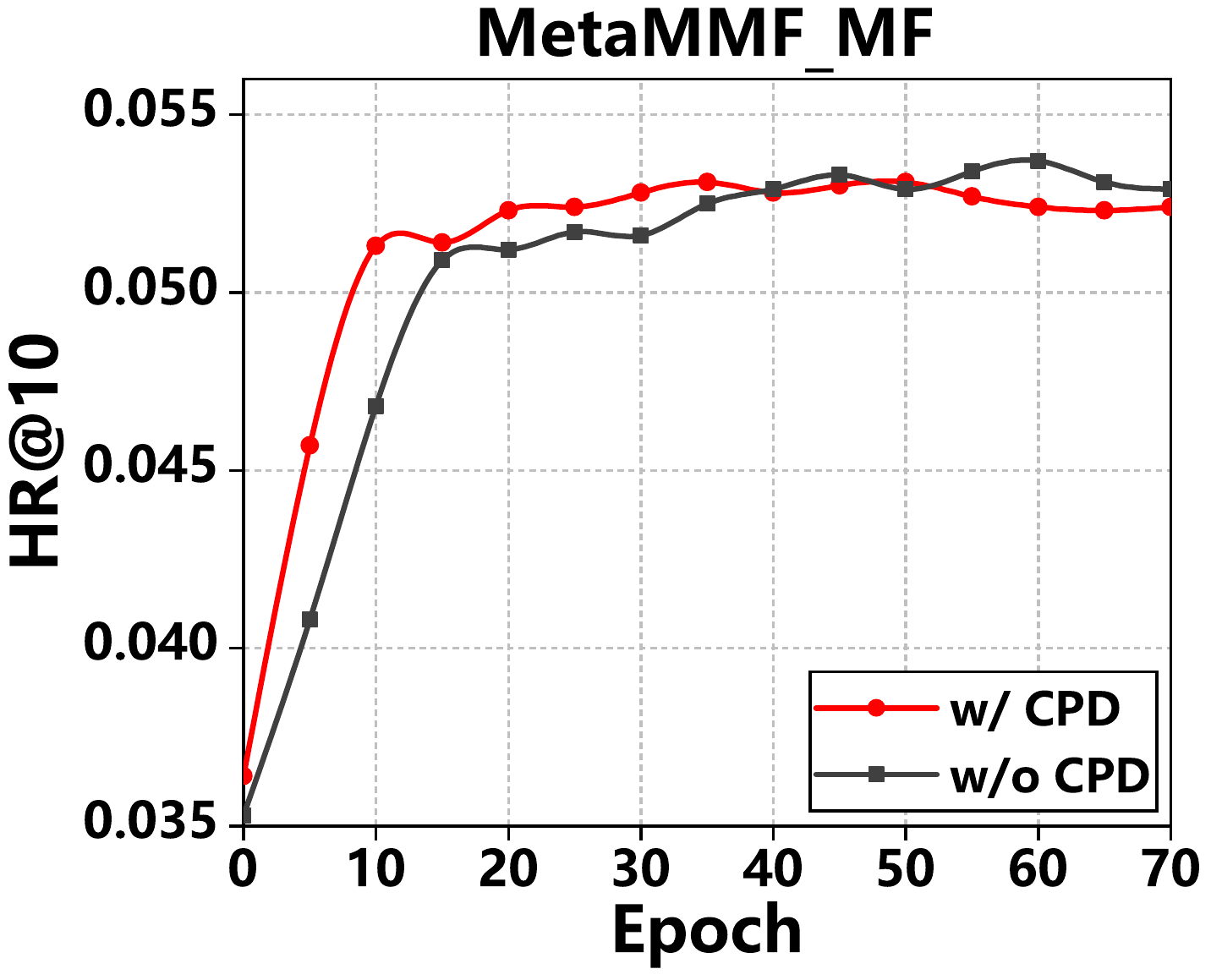}
\end{minipage}
\hspace{1cm}
\begin{minipage}[t]{0.33\linewidth}
\centering
\includegraphics[width=\linewidth]{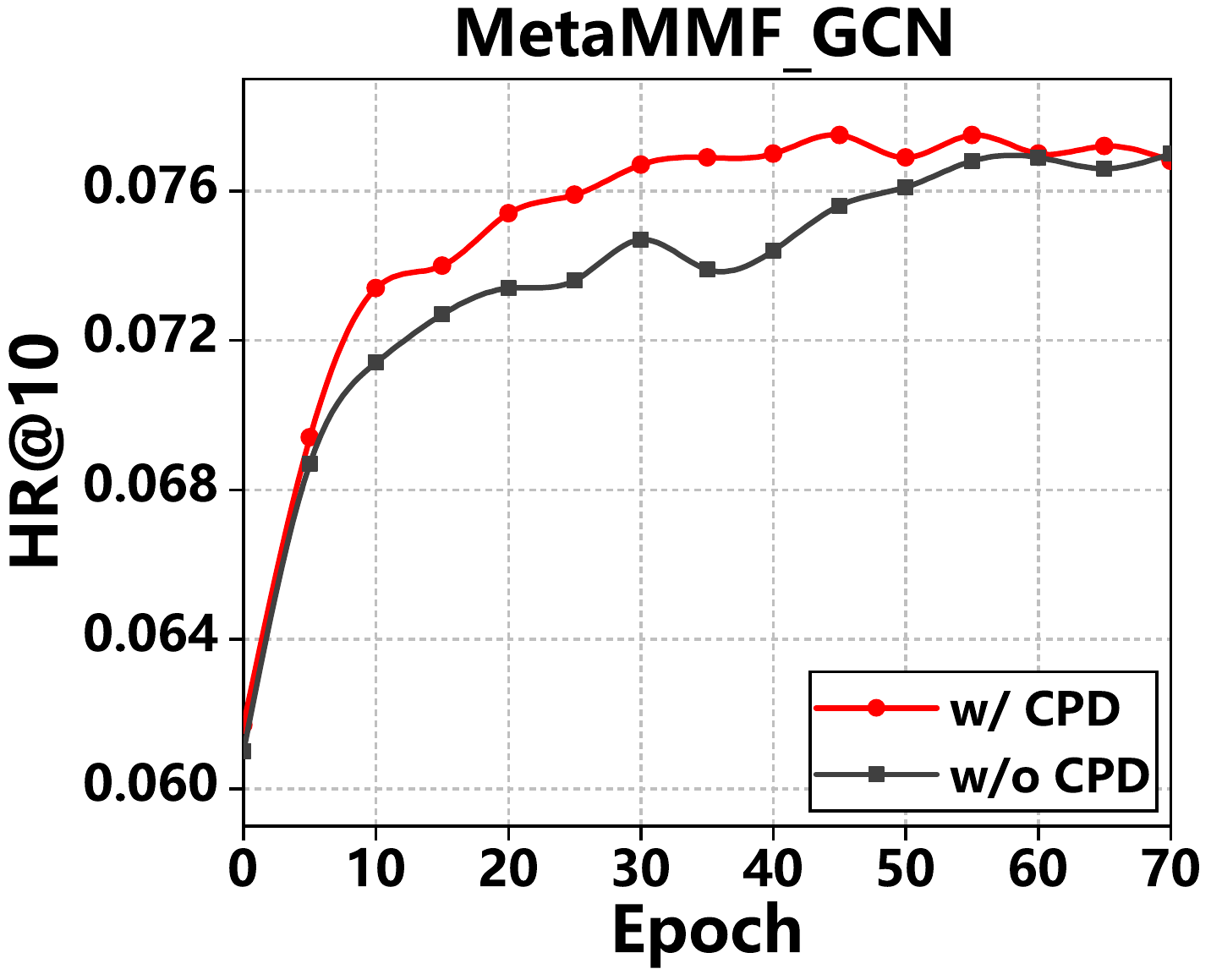}
\end{minipage}
\centerline{TikTok}
\\
\begin{minipage}[t]{0.33\linewidth}
\centering
\includegraphics[width=\linewidth]{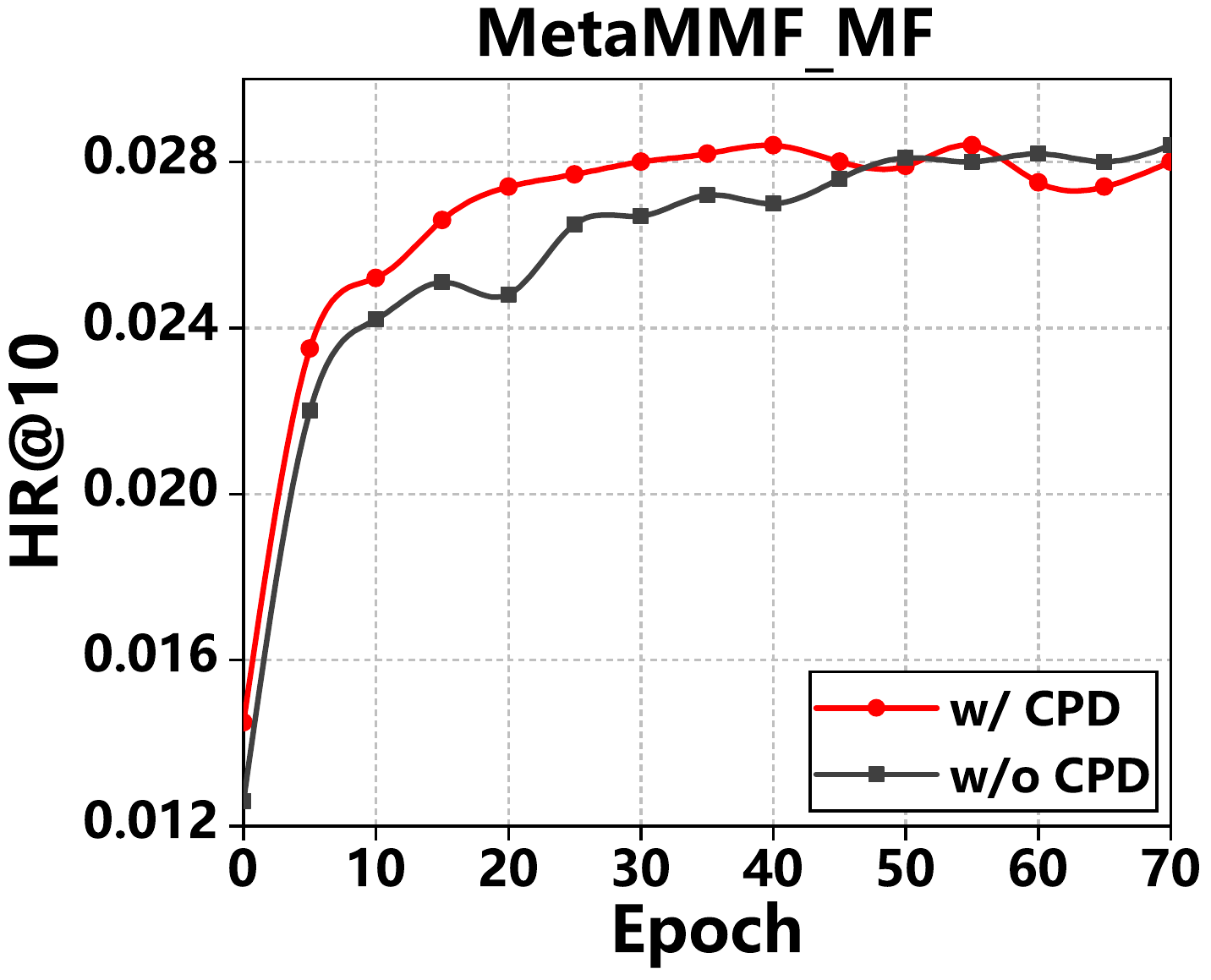}
\end{minipage}
\hspace{1cm}
\begin{minipage}[t]{0.33\linewidth}
\centering
\includegraphics[width=\linewidth]{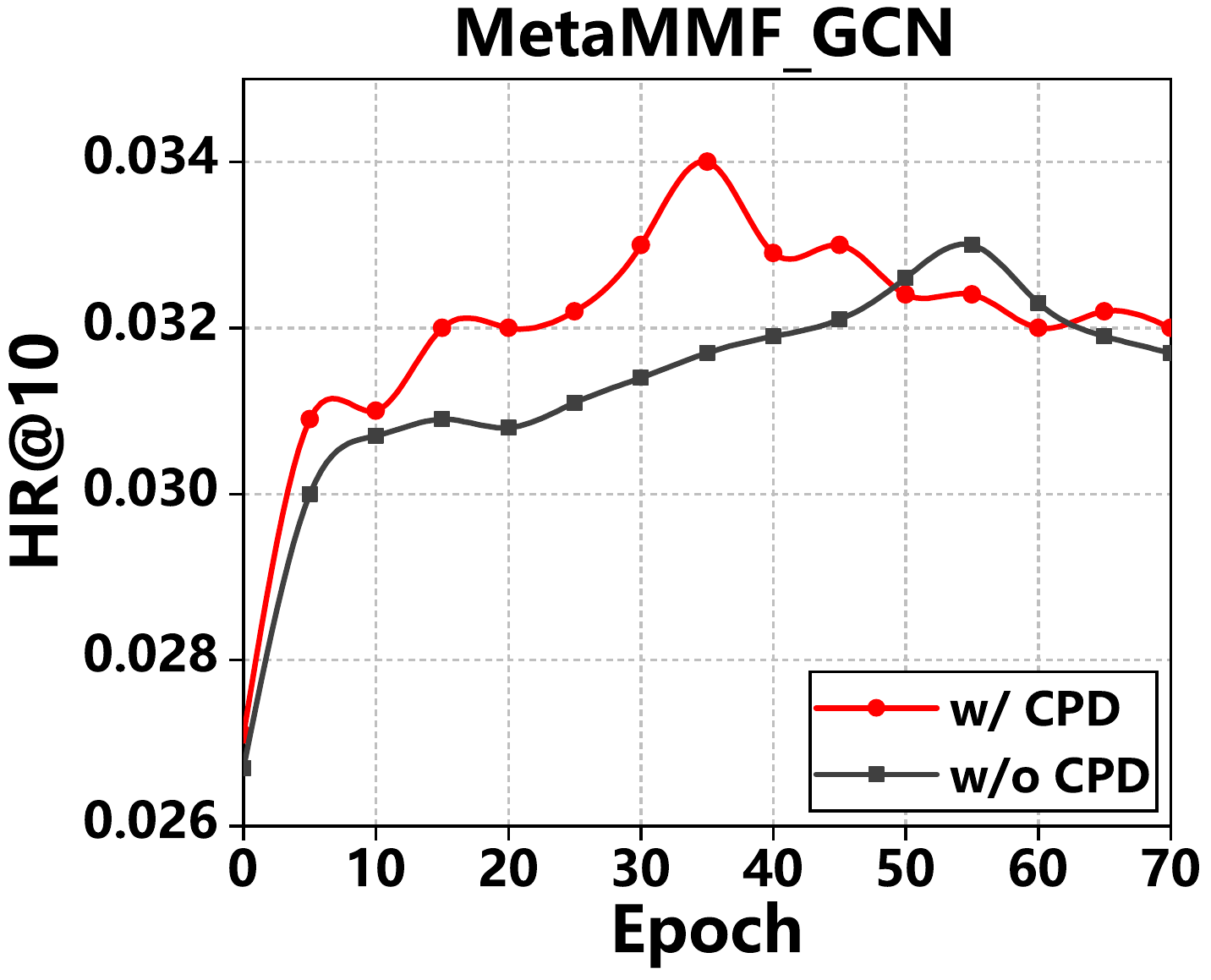}
\end{minipage}
\centerline{Kwai}
\centering
\vspace{-0.4cm}
\caption{Test performance of each epoch of MetaMMF\_MF and MetaMMF\_GCN with and without CP decomposition.}
\label{fig:epoch}
\vspace{-0.3cm}
\end{figure}

Considering that CP decomposition effectively reduces the parameters in MetaMMF, we can infer that the simplified model can be trained to converge more easily. To validate this inference, we conducted experiments to explore how CP decomposition affects the convergence rate of the model. Figure~\ref{fig:epoch} shows the test performance \textit{w.r.t.} HR@$10$ of MetaMMF\_MF and MetaMMF\_GCN with (w/) and without (w/o) CP decomposition as the training epoch progressing. Due to the space limitation, we omitted the performance \textit{w.r.t.} other evaluation metrics which has a similar trend. As shown in Figure~\ref{fig:epoch}, we can see that:

\begin{itemize}[leftmargin=*]
\item On three datasets, the two methods exhibit faster convergence with CP decomposition compared with using the complete 3-D tensor parameters. The gap of convergence rate caused by CP decomposition is more obvious on the MovieLens dataset where MetaMMF is equipped with more fusion layers. This phenomenon is reasonable since CP decomposition can substantially reduce the number of to-be-optimized parameters and thus improve the efficiency of model training.
\item Besides, CP decomposition slightly influences the performance of MetaMMF\_MF while enhancing the performance of MetaMMF\_GCN. The reason might be that the complete MetaMMF is more likely to suffer from over-fitting when combined with a complex GCN model, and CP decomposition resolves this issue by simplifying the model complexity. This demonstrates that CP decomposition benefits MetaMMF in multiple aspects, including size, convergence, and model performance.
\end{itemize}

To demonstrate the time efficiency resulting from CP decomposition transparently, we measured the training time of two models, MetaMMF\_MF and MetaMMF\_GCN, both with (w/) and without (w/o) CP decomposition. For a fair comparison, all the models were trained on Ubuntu 16.04.5 with NVIDIA TITAN Xp and Python3.7. Table~\ref{tab:training-time} displays the range of training epochs (eps) when the various variant models achieve convergence. Additionally, we recorded the training time in minutes taken by the variants to complete the unit training epoch (min/ep). From Table~\ref{tab:training-time}, it is evident that the decomposed models converge in fewer epochs and require less training time to complete one training epoch. Therefore, we conclude that CP decomposition decreases training time not only by earlier convergence but also by reducing the time taken per epoch. The reason is that CP decomposition effectively reduces the number of parameters that must be optimized.

\begin{table}[t]\centering
\caption{Training time comparison of MetaMMF with and without CP decomposition.}
\label{tab:training-time}
\vspace{-0.3cm}
\setlength{\tabcolsep}{1mm}{
\begin{tabular}{|l|c|cc|cc|cc|}
\hline
\multirow{3}*{Methods} & \multirow{3}*{CPD} & \multicolumn{2}{c|}{MovieLens} & \multicolumn{2}{c|}{TikTok} & \multicolumn{2}{c|}{Kwai} \\
&  & \small{Converge} &  \small{Training time} &  \small{Converge} &  \small{Training time} &  \small{Converge} &  \small{Training time} \\
&  & (eps) &  (min/ep) &  (eps) &  (min/ep) &  (eps) &  (min/ep) \\
\hline
\multirow{2}*{MetaMMF\_MF} & w/ & 45$\sim$50 & 6.07 & 30$\sim$35 & 2.33 & 35$\sim$40 & 0.88 \\
& w/o & 65$\sim$70 & 8.74 & 55$\sim$60 & 2.62 & 50$\sim$55 & 1.05\\
\hline
\multirow{2}*{MetaMMF\_GCN} & w/ & 40$\sim$45 & 11.69 & 40$\sim$45 & 5.08 & 30$\sim$35 & 1.42\\
& w/o & 55$\sim$60 & 12.41 & 55$\sim$60 & 5.32 & 50$\sim$55 & 1.47\\
\hline
\end{tabular}}
\vspace{-0.3cm}
\end{table}

\subsection{Effect of Static Weights (RQ5)}
Additionally, we conducted ablation experiments to investigate the impact of static weights on our dynamic model. Table~\ref{tab:ablation_static_weight} presents the performance of our two methods, with (w/) and without (w/o) static weights, in terms of HR@10 and NDCG@10 on three datasets. The results indicate that the performance of MetaMMF is negatively affected when the static weights are eliminated. This observation emphasizes the importance of shared and static fusion parameters in our dynamic fusion model. To figure out why, we think the static parameters serve as fundamental and universal multimodal fusion components for all micro-videos. From a meta-learning perspective, the static component can be viewed as learning the initialization parameters to generalize most tasks, which lays the groundwork for subsequent fine-tuning to adapt to each unique task.

\begin{table}[t]\centering
\caption{Performance comparison of MetaMMF with and without static weights.}
\label{tab:ablation_static_weight}
\vspace{-0.3cm}
\begin{tabular}{|l|c|cc|cc|cc|}
\hline
\multirow{2}*{Methods} & Static & \multicolumn{2}{c|}{MovieLens} & \multicolumn{2}{c|}{TikTok} & \multicolumn{2}{c|}{Kwai} \\
& weights & HR@10 &  N@10 &  HR@10 &  N@10 &  HR@10 &  N@10 \\
\hline
\multirow{2}*{MetaMMF\_MF} & w/ & 0.1539 & 0.1268 & 0.0521 & 0.0422 & 0.0285 & 0.0336 \\
& w/o & 0.1460 & 0.0987 & 0.0520 & 0.0403 & 0.0273 & 0.0324\\
\hline
\multirow{2}*{MetaMMF\_GCN} & w/ & 0.2018 & 0.1757 & 0.0775 & 0.0652 & 0.0340 & 0.0398\\
& w/o & 0.2016 & 0.1753 & 0.0769 & 0.0638 & 0.0312 & 0.0370\\
\hline
\end{tabular}
\vspace{-0.4cm}
\end{table}
\section{Conclusion and Future Work}
In this work, we upgraded the multimodal fusion from static to dynamic to enhance the representation learning in the micro-video recommendation. We devised a novel meta-learning model MetaMMF, which realizes the target by learning the item-specific multimodal fusion functions for different items. The core of MetaMMF is the meta fusion learner which can adaptively parameterize a neural network for an item's multimodal fusion based on the meta information derived from its multimodal features. Extensive experiments on three real-world datasets demonstrate the rationality and effectiveness of leveraging dynamic fusion to learn multimodal representations. In the future, we will exploit MetaMMF to dynamically integrate user preferences on multiple modalities and produce higher-quality user representations, which will be beneficial to recommendation performance. Moreover, we are interested in introducing MetaMMF to address the cold-start problem of multimodal recommendation~\cite{du2020learn}.

This work is an initial attempt at dynamic multimodal fusion, limited to a neural network framework and the micro-video recommendation scenario. Specifically, there are many other types of fusion functions, such as multiple kernel learning methods~\cite{gonen2011multiple} and graphical models. Furthermore, multimodal fusion is widely applied in various tasks, such as emotion recognition~\cite{cowie2001emotion}, multimedia event detection~\cite{lan2014multimedia}, and multimedia retrieval. We expect the potential of dynamic multimodal fusion can be further explored towards other model structures and tasks.


\begin{acks}
This work is supported by the National Natural Science Foundation of China, No.:U1936203; and the program of China Scholarship Council, No.:202106220122.
\end{acks}

\bibliographystyle{ACM-Reference-Format}
\bibliography{samples/ref}

\appendix









\end{document}